%% file: main.tex
\documentclass[manuscript,screen,nonacm]{acmart}

\renewcommand\footnotetextcopyrightpermission[1]{}
\setcopyright{none}

\usepackage{dsfont}
\usepackage{booktabs}
\usepackage{multirow}
\usepackage{tikz}
\usepackage{algorithm}
\usepackage{algpseudocode}

\theoremstyle{acmdefinition}
\newtheorem{assumption}{Assumption}
\newtheorem{remark}{Remark}

\graphicspath{{figures/}}

\newcommand{\pr}{\mathbb{P}}
\newcommand{\ind}{\mathds{1}}
\newcommand{\lcb}{\underline{p}}          
\newcommand{\lcbk}{\underline{\kappa}}    
\newcommand{\xadj}{\vec{x}_{\mathrm{adj}}}      
\newcommand{\xp}{\vec{x}_p}               
\newcommand{\Xother}{X_o}                 
\newcommand{\xo}{\vec{x}_o}               

\begin{document}

\title{Now You See Me Now You Don't: A Framework To Detect and Reduce Proxy Capacity}

\author{Nirek Sharma}
\authornote{The views expressed are the author's own and do not necessarily reflect those of Upstart Network, Inc., its affiliates, or its employees. This paper presents general research on methodology. It is not legal advice and should not be read as describing whether or how Upstart or its lending partners use these methods in any model, product, or credit decision. See Competing Interests.}
\affiliation{%
  \institution{Upstart Network, Inc.}
  \country{USA}}
\email{nirek.sharma@upstart.com}

\begin{abstract}
  \input{sections/00_abstract}
\end{abstract}



\maketitle

\input{sections/01_introduction}
\input{sections/02_proxy_test}
\input{sections/03_feature_adjuster}

\input{sections/04_synthetic}
\input{sections/05_real_world}
\input{sections/06_discussion}
\input{sections/08_conclusion}

\clearpage
\input{sections/99_endmatter}

\bibliographystyle{ACM-Reference-Format}
\bibliography{refs}

\appendix
\renewcommand{\thesection}{Appendix~\Alph{section}}
\input{sections/A_uncertain_z}
\input{sections/A2_estimate_uncertainty}

\input{sections/B_synthetic_archetypes}
\input{sections/D_real_world_audit}
\input{sections/E_causal_proxy}

\end{document}

%% file: sections/00_abstract.tex

In regulated domains such as lending, the two most salient fairness concepts are disparate impact and disparate treatment: disparate impact is generally defined by differential rates of adverse outcomes between demographic groups, while disparate treatment is generally defined by the use of demographic information or proxies to issue decisions.
There already exist established methods to quantitatively detect and mitigate disparate impact; however, there has been less research on doing the same for disparate treatment.
From a practitioner's perspective, the manner in which a conditional independence test is formulated and operationalized to identify proxies is nuanced and underexplored.
As Machine Learning (ML) systems increase in scope and application in regulated domains, practitioners need a defensible framework to identify proxies at scale.
Current disparate treatment frameworks typically rely upon some combination of quantitative and qualitative review of models' inputs to identify proxies.
Features that are identified as proxies are then typically removed from the model in practice, and predictive performance degrades in favor of fairness through unawareness.
Here, we propose a two-part framework that reduces the disparate treatment risk of a feature and keeps the predictive performance that removal of the feature loses.
First, we propose a quantitative test for proxy capacity that identifies whether some of a feature's values reveal membership in a particular demographic group.
Second, we propose a pre-processing methodology that adjusts those proxy features to reduce their proxy capacity.
We show empirically on synthetic and public datasets that this framework identifies features that proxy for demographic information and reduces their proxy capacity while the adjusted feature retains task-relevant signal (i.e.\ improves predictive performance versus removal).

%% file: sections/01_introduction.tex
\newcommand\independent{\protect\mathpalette{\protect\independenT}{\perp}}
\def\independenT#1#2{\mathrel{\rlap{$#1#2$}\mkern2mu{#1#2}}}

\section{Introduction}
\label{sec:intro}

Artificial intelligence systems that drive decisions in regulated domains such as lending are generally subject to anti-discrimination law that recognizes two theories of harm \citep{taylor2018ecoa}: \emph{disparate treatment}, in which protected attributes such as race or sex are used to issue decisions \citep{ricci2009}, and \emph{disparate impact}, in which a facially neutral decision process produces disproportionate outcomes across demographic groups \citep{griggs1971}.

A robust body of literature has helped advise practitioners on how to approach detecting \citep{feldman2015certifying, hardt2016equality, corbett2017algorithmic, kleinberg2017inherent, chouldechova2017fair, pleiss2017fairness, corbett2023measure} and mitigating \citep{kamiran2012data, zemel2013learning, zafar2017fairness, agarwal2018reductions, mehrabi2021survey, barocas2023fairness} disparate impact (or indirect discrimination). In the context of lending, for example, practitioners calculate Adverse Impact Ratios (AIRs) to establish if disparate impact exists \citep{feldman2015certifying, black2026fairlending}.
Subsequently, a variety of methods exist that can be used to find a Less Discriminatory Alternative (LDA) model \citep{gillis2024operationalizing, black2026fairlending} such as adversarial debiasing \citep{zhang2018mitigating} or post-processing \citep{hardt2016equality, tifrea2024frappe}. However, it has been argued that LDAs need not be adopted if they fail to satisfy the ``business objective'' however defined by the practitioner \citep{weerts2024unlawful, black2026fairlending}.
Continuing with our lending example: if two groups are observed to have systematically different distributions of creditworthiness, then it has been argued that an LDA that deteriorates accuracy in favor of improved AIRs may fail to satisfy the business objective and be declared inviable \citep{corbett2017algorithmic, kleinberg2017inherent}.

Therefore, recent work has called for focus on disparate treatment, or direct discrimination \citep{tschantz2022proxy, adamsprassl2023directly, binns2023legal, weerts2024unlawful}.
Minimizing disparate treatment is typically done by building models without the use of protected attributes: fairness through blindness or unawareness. The historical precedent of this is clear \citep{taylor2018ecoa, balkin2003anticlassification}; however, blindness only to protected attributes is insufficient for machine-learning-based solutions since ostensibly neutral features such as ZIP codes, surnames, or
occupations can carry the same information as the protected attribute and a model trained
using those features can be argued to present disparate treatment
\citep{barocas2016big, prince2020proxy}.

\citet{kumar2022equalizing} describe identifying proxies through a feature's association with a demographic group, and cite the US Federal Reserve's definition \citep{avery2012credit} requiring that the feature's predictiveness of the target arise only through that association (see the causal graph provided in~\ref{app:causal-proxy}).
Practically speaking, however, this is insufficient since this definition is inherently target-dependent: the test requires labels $Y$, so a feature's proxiness is tied to a particular prediction task rather than to the feature itself, and it ignores the feature's overall relationship to group membership.
\citet{weerts2024unlawful} further separate \emph{simple} proxies, a single feature linked to the protected characteristic, from \emph{complex} proxies, a set of features that are linked to it together.
Thus, since fairness through blindness to \emph{just} protected characteristics is insufficient, practitioners need a sensible way of identifying features that are strongly related to (or proxy for) demographic information.
\citet{black2026fairlending} found that practitioners frequently rely on a combination of qualitative review and ad hoc quantitative review.
The quantitative review looks for proxies that are ``highly correlated with a prohibited basis,'' and institutions often keep ``green and yellow and red lists for variables.''
Previous literature on how to quantitatively identify simple proxy features leaves two open questions:

\begin{enumerate}
  \item It is not clear how best to measure the relationship between a feature and protected characteristics. It has been previously suggested to use aggregate measures of association such as correlation or mutual information (MI) \citep{weerts2024unlawful, black2026fairlending}; however, \citet{weerts2024unlawful} identified that such measures may miss associations that only hold for part of a demographic group.
  \item Further, it is unclear how to operationalize such a measure of association. Practitioners often gravitate towards ``Low/Medium/High'' classifications for proxy features, but it is not clear how to set thresholds that are broadly applicable across demographic groups. In \citet{black2026fairlending}, it was noted that there are ``still no metrics or no concrete, crispy, guidance'' from regulators.
\end{enumerate}

\paragraph{Our Contributions}
In this paper, we aim to make two key contributions:

\begin{enumerate}
  \item We propose a novel framework to identify proxy features. This framework allows practitioners to apply a consistent criterion in identifying whether features input into ML system have significant proxy capacity regardless of the type of feature, shape of relationship between the feature and demographic group, or the rarity of the demographic group. This allows practitioners to increase the scale oversight of ML systems in regulated domains by applying a consistent criterion to increasingly wider datasets.
  \item We propose a novel methodology to generally reduce this proxy capacity of a feature as an alternative to outright removal or unmodified inclusion. Practitioners are faced with a binary decision that either deteriorates model performance or accepts some regulatory risk respectively. This methodology offers a novel third option which can balance the costs of both options.
\end{enumerate}

\begin{figure}[t]
  \centering
  \includegraphics[width=0.7\textwidth]{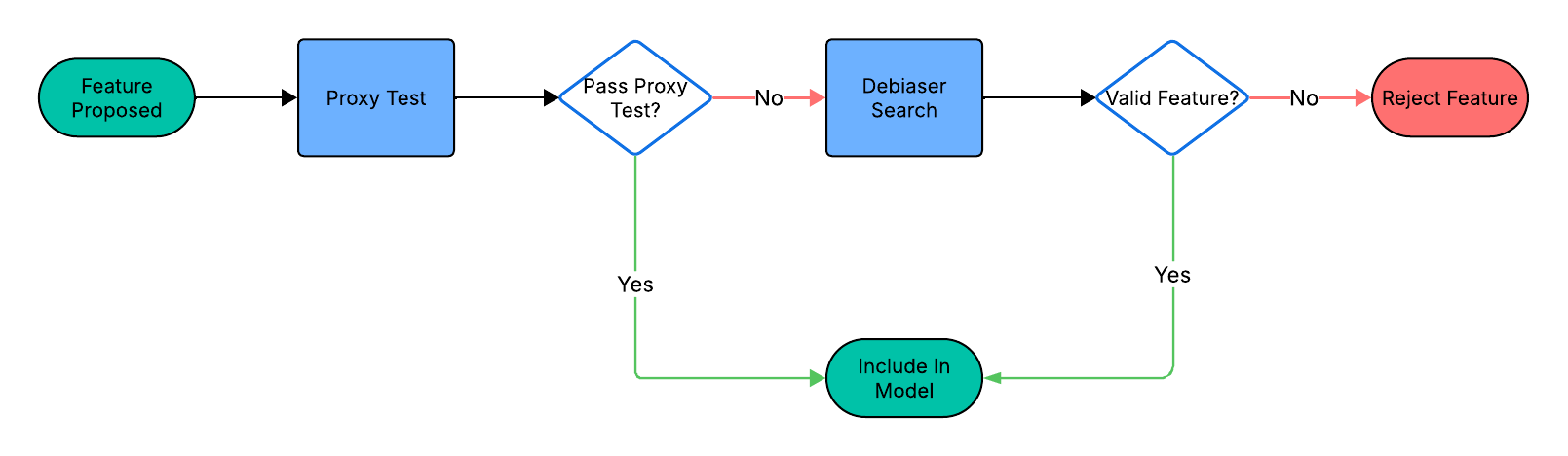}
  \Description{Flow chart. A proposed feature goes to the proxy test. If it passes, it is included in the model. If it fails, it goes to the debiaser search; if the search yields a valid feature, it is included in the model, and otherwise the feature is rejected.}
  \caption{
    The proposed feature governance workflow: features that are identified as proxies by the \emph{distributional proxy test} (Section~\ref{sec:proxy-test}) are sent to the Debiaser Search (Section~\ref{sec:adjuster}) to find an adjuster model that yields a valid, non-proxy adjusted feature that still improves model accuracy.
  }
  \label{fig:feature_workflow}
\end{figure}

The effect of this paper is an operationalization of \citet{weerts2024unlawful} through a framework that first identifies features with significant \emph{simple proxy capacity} and then mitigates that capacity by adjusting the inputs (Figure~\ref{fig:feature_workflow}).

In Section~\ref{sec:proxy-test}, we introduce a \emph{distributional proxy test} that identifies features with a significant proxy capacity (Definition~\ref{def:local-capacity}) based on estimating $\pr(Z{=}1 \mid X)$ for demographic group $Z$ and feature $X$ and comparing it to a carefully chosen threshold $T$.
The test has four properties that make it suitable for feature governance:
\begin{itemize}
  \item \textbf{Local.}
        The test flags a feature when any region of $X$ concentrates group membership, whatever the shape of $\pr(Z{=}1 \mid X)$. The exact method of estimation is a choice for the practitioner; in this paper we illustrate one such method of estimating $\pr(Z{=}1 \mid X)$ which does not assume a distribution of $X$.
        We show in Section~\ref{sec:exp-synthetic} how aggregate measures like MI struggle to identify features with local pockets of group membership which are identified with the proposed methodology.
  \item \textbf{Base-rate aware.}
        We propose to derive thresholds per demographic group so the same strength of evidence is required to identify proxies for all demographic groups.
  \item \textbf{Flexible.}
        We allow for practitioners to easily parameterize what strength of association is required for a feature to be identified as a proxy to reflect the organization's position and domain expertise. This critically allows practitioners to distinguish between ``statistical'' and ``practical'' significance of the relationship between the feature and demographic group membership. This can easily be extended to give a principled intuition on classifying features as Low, Medium, and High risk (Section~\ref{sec:thresholds}).
  \item \textbf{Calibrated for imputed demographics.}
        Protected class labels are rarely observed directly and are usually imputed in lending \citep{cfpb2014bisg, elliott2009bisg}, which can bias, and add uncertainty to, estimates of group disparities \citep{chen2019fairness, kallus2022assessing, elzayn2023estimating}.
        The proposed framework allows practitioners to calibrate the test's thresholds to account for this uncertainty (\ref{app:uncertain-z}).
\end{itemize}

In Section~\ref{sec:adjuster}, we present a \emph{feature adjuster} that pre-processes proxy features to reduce their simple proxy capacity.
Like some LDA search methodologies (such as adversarial debiasing), this procedure utilizes the ability to recover group membership in the aggregate through a set of non-proxy features in order to minimize a feature's simple proxy capacity \citep{lipton2018does, black2026fairlending}.
This effectively allows practitioners to ensure that no single feature concentrates demographic information beyond what is already present in the data.
\begin{itemize}
  \item The adjuster yields a per-record correction learned from the model's other, non-proxy features, trained against a differentiable approximation of the \emph{distributional proxy test}.
  \item $Z$ is used only to identify the proxies and to train the adjuster: at inference the downstream model only has access to non-proxy features.
  \item The function class of the adjuster is flexible: we demonstrate a multi-layer perceptron (MLP) and a gradient-boosted tree ensemble (XGBoost).
\end{itemize}

We validate the framework on synthetic features with known proxy capacity, where the test identifies proxies that aggregate measures miss and the adjuster reduces their proxy capacity while retaining predictive value over removal (Section~\ref{sec:synthetic}), and we also replicate these results on a public dataset (Section~\ref{sec:real-world}).

%% file: sections/02_proxy_test.tex
\section{Distributional Proxy Test for Proxy Capacity}\label{sec:proxy-test}

Let us first define the terms useful to formulating our proposed test for proxy capacity.
In a dataset of $n$ records, we treat a single feature, $X$, as a random variable that takes values $x$.
Let $Z \in \{0,1\}$ represent the protected attribute indicating membership in the demographic group being tested.
We refer to the base rate of group membership as $\pi = \pr(Z{=}1)$.
We consider one demographic group at a time, so all quantities in this paper, including $\pi$, are specific to the group under test.
Often $Z$ is not observed directly; in its place we have a probabilistic estimate $\hat Z \in [0,1]$.\footnote{In the case of lending, practitioners often rely upon Bayesian Improved Surname Geocoding (BISG) and Bayesian Improved First Name Surname Geocoding (BIFSG) to produce probabilities of group membership \citep{elliott2009bisg, voicu2018bifsg}.}
Using these terms, we formalize the definition of \emph{proxy capacity} given by \citet{weerts2024unlawful} below:

\begin{definition}[Proxy Capacity]\label{def:capacity}
  \emph{Proxy capacity} is the extent to which the protected attribute $Z$ can be recreated from the feature $X$ \citep{weerts2024unlawful}.
\end{definition}

\begin{definition}[Local Proxy Capacity]\label{def:local-capacity}
  The \emph{local proxy capacity} of some region of $X$ is the strength of evidence it carries for group membership.
  A feature's proxy capacity, as considered here, is the largest local capacity over all its regions (Section~\ref{sec:estimating}).
\end{definition}

\begin{remark}[Why Test Local Capacity]\label{rem:leakage}
  Group information can be concentrated in a single region or value of $X$ even when the feature reveals little about the group overall.
  \citet{weerts2024unlawful} provide this motivation in their Section 4.1.
  For example, imagine a binary variable that is 0 for all non-members, and 1 for some members of $Z$.
  Sufficiently complex non-linear models can separate on exactly such values, so one revealing value or region is enough for a model to treat a subgroup differently, which is precisely the disparate treatment risk we aim to identify.
  The test therefore measures a feature's most revealing value or region of values rather than an average.
\end{remark}

We operationalize these definitions through the \emph{distributional proxy test} given by Equation~\eqref{eq:z_given_x}:

\begin{equation}\label{eq:z_given_x}
  \max_X (\pr(Z{=}1 \mid X)) \ge T
\end{equation}
In words, the test identifies a feature as a proxy when some region of $X$ has a rate of group membership of at least $T$.
A region is a set of values of $X$, and the practitioner chooses how small a region can be (Section~\ref{sec:estimating}) and the threshold $T$ (Section~\ref{sec:thresholds}).

\subsection{Estimating \texorpdfstring{$\pr(Z{=}1 \mid X)$}{P(Z=1 | X)}}\label{sec:estimating}

Equation~\eqref{eq:z_given_x} relies upon estimating the distribution of group membership $Z$ given the observed distribution of $X$.
For categorical or binary features, the practitioner can simply calculate $\pr(Z = 1 \mid X = x)$ for all observed values of $x \in X$.
For those features, calculating Definition~\ref{def:local-capacity} is straightforward: the proxy capacity of $X$ is simply the maximum observed rate of group membership over the support of $X$: $\max_{x \in X}\pr(Z{=}1 \mid X = x)$.\footnote{Of course, practitioners need to take care to calculate the appropriate support for infrequent values: infrequent categories will provide noisy estimates.}

For numeric features, we need a sensible way of estimating the same quantity over a continuous distribution of $X$.
For a region $R$ of $X$ with density $f$, and $m(x) = \pr(Z{=}1 \mid X{=}x)$, the rate of group membership in $R$ is the density-weighted average of $m$ over $R$:
\begin{equation}\label{eq:integral}
  \pr(Z{=}1 \mid X \in R) \;=\; \frac{\int_R m(x)\, f(x)\, dx}{\int_R f(x)\, dx}.
\end{equation}
There are many possible ways of estimating this quantity practically, and this can be treated as an implementation choice for the practitioner.
For illustrative purposes, we propose to simply segment $X$ into regions $R$ over which $\pr(Z{=}1 \mid X \in R)$ can be estimated; however, the granularity of the regions dictates the usefulness of the estimate.
At one extreme, one region holds every record, which gives the overall rate of group membership, $\pi$.
At the other extreme, each region holds one record, so its rate is $Z_i$, and the maximum rate of group membership is simply $1$.
Neither extreme is useful, so the practitioner must choose a granularity that balances the probability of Type 1 and Type 2 errors.
Let $q$ be the share of records in $R$: a smaller $q$ finds smaller pockets of group membership but identifies more features as proxies by chance, and a larger $q$ does the reverse.
One simple way to partition the observed values of $X$ is to use its empirical quantiles: splitting $X$ into $n_{\text{bins}} = \frac{1}{q}$ regions.\footnote{
  Bins have fixed boundaries, so a pocket of group membership that straddles a boundary is split between two bins and diluted in both.
  Using \emph{rolling windows} avoids this: each window is a run of consecutive records, sorted by $X$, that holds a share $q$ of the records. But this tests $\sim n$ overlapping regions, requiring stronger correction for multiple comparisons.
}

\paragraph{Estimating the rate of group membership within a region.}
For each region $R$, let $n_R$ be the number of records in the region and $s_R = \sum_{i:\, X_i \in R} Z_i$ the number of group members.\footnote{In the cases where $Z$ is not observed, we replace $Z_i$ with its estimate $\hat Z_i$ to get the \emph{expected} number of group members in $R$.}
A simple average estimates the rate of group membership in $R$ as its share of group members, $\hat p_R$, and the estimate of the feature's proxy capacity is the largest such share:
\begin{equation}\label{eq:binned}
  \hat p_R \;=\; \frac{s_R}{n_R},
  \qquad
  \hat\kappa \;=\; \max_R \hat p_R .
\end{equation}
To decrease noise in small regions, a practitioner can instead use a smooth estimate $\hat m(X_i)$ of $m(X_i)$ over the records in each region, such as a logistic spline for example \citep{eilers1996flexible}.

\paragraph{Considerations when estimating $\pr(Z{=}1 \mid X)$}
We suggest that practitioners compute a confidence interval about this point estimate and take care to avoid multiple hypothesis testing errors, since $\hat\kappa$ is a maximum over regions.\footnote{\ref{app:estimate-uncertainty} describes how we calculate a confidence interval about this in our experiments.}
The test identifies a feature as a proxy when its estimate of proxy capacity (Definition~\ref{def:local-capacity}) reaches $T$, that is, when some region of $X$ indicates membership in group $Z$.
Practitioners can incorporate domain knowledge in their choice of granularity $q$, in the same way as in their choice of threshold described below (Section~\ref{sec:thresholds}).

\subsection{Choosing \texorpdfstring{$T$}{T}}\label{sec:thresholds}

In order to test for the same local proxy capacity for groups with differing base rates $\pi$, we calculate the threshold $T$ through Bayes' rule.
The posterior odds that a record in region $R$ belongs to group $Z$ are the evidence the region provides, or the \emph{Bayes factor} $B$, times the prior odds of membership:
\begin{equation}
  \underbrace{\frac{\pr(Z{=}1 \mid X \in R)}{\pr(Z{=}0 \mid X \in R)}}_{\text{posterior odds}}
  \;=\;
  \underbrace{\frac{\pr(X \in R \mid Z{=}1)}{\pr(X \in R \mid Z{=}0)}}_{\text{Bayes factor}}
  \cdot
  \underbrace{\frac{\pi}{1-\pi}}_{\text{prior odds of membership}}
\end{equation}
Consider an illustrative example where we used a fixed threshold on the posterior probability, such as $\pr(Z{=}1 \mid X \in R) \ge 0.95$.
For a group with an even base rate ($\pi = 0.5$), the prior odds are 1, so the rule demands the strength of evidence $B \geq 0.95/(1 - 0.95) = 19$.
Now consider a rare group at $\pi = 0.03$, the same rule demands $B \approx 614$, but for a more frequent (majority) group at $\pi = 0.8$, it demands only $B = 4.75$.
Clearly, it would be more difficult for rare demographic groups to provide sufficient evidence to identify proxy features.

Instead of holding the posterior constant, we propose to fix the strength of evidence $B$ for every group and calculate the threshold as
\begin{equation}\label{eq:thresh}
  T(B, \pi) \;=\; \frac{B\pi}{1 - \pi + B\pi}.
\end{equation}
Intuitively, $T(B, \pi)$ is the rate of group membership a region reaches when evidence of strength $B$ acts on a prior $\pi$.
Inverting Equation~\eqref{eq:thresh} gives the \emph{implied Bayes factor} of an observed statistic, $B(\hat\kappa, \pi) = \frac{\hat\kappa}{1 - \hat\kappa} \cdot \frac{1 - \pi}{\pi}$, so every result can be read at any $B$.

When $Z$ is not observed, the test uses probabilistic estimates $\hat Z$ such as BISG and BIFSG, and these estimates bias $\hat\kappa$.
For a calibrated estimator, the estimate of a region's rate is pulled toward the base rate $\pi$, so a region that truly reaches $T$ appears to fall below it.
We therefore suggest that practitioners lower the threshold to the value that a true proxy is expected to reach.
\ref{app:uncertain-z} shows how to adjust the threshold $T$ to account for this bias given a probabilistic demographic estimator.

\paragraph{Fairness posture.}
As described, the choice of $B$ should be set by the practitioner, and it should reflect the organization's position and its domain expertise.
Domain expertise matters because some features carry evidence of group membership and have nonetheless been reviewed qualitatively and agreed to be acceptable.
Credit scores, for example, clearly differ between demographic groups \citep{avery2012credit}, yet nonetheless it has been agreed upon that those features do not present disparate treatment risk \citep{black2026fairlending}.
The implied Bayes factors of such agreed-upon features can provide practitioners natural reference points when setting $B$.
Further, practitioners can set multiple levels $B_{\mathrm{low}} < B_{\mathrm{high}}$ to yield three risk tiers based on the implied Bayes factor: High ($\ge B_{\mathrm{high}}$), Medium, and Low ($< B_{\mathrm{low}}$) \citep{black2026fairlending}.
Practitioners can refer to the literature on Bayes factors for guidance on how to interpret and set $B$ \citep{jeffreys1961theory, kass1995bayes}.

%% file: sections/03_feature_adjuster.tex
\section{Feature Adjuster For Minimizing Proxy Capacity}\label{sec:adjuster}

Features that have been identified as proxies are typically removed outright from the model.
However, one could argue that features can instead be processed in such a way to maintain their predictive power while limiting their proxy capacity.
Consider an hypothetical illustration, again in the context of lending: let one possible feature be the \emph{length of credit history}.
One could certainly find that this feature would be valuable to an underwriting model, but also could be found to be a proxy for \emph{age}, a prohibited basis \citep{avery2012credit}.
What a practitioner could do is simply trim the value of that feature, creating an adjusted feature that no longer proxies for age but still provides some signal vs. removal.
The proposed methodology in this section generalizes this approach to any arbitrary proxy feature.

Let us denote $\xp \in \mathbb{R}^n$ as a feature that proxies for protected attribute $Z$.
Let us also denote \emph{other} (non-proxy) features as $\Xother \in \mathbb{R}^{n \times d}$ with individual records $\vec{x}_{o} \in \mathbb{R}^d$.
We aim to find an \emph{adjuster} function $g : \mathbb{R}^d \to \mathbb{R}$ that takes a record's non-proxy features $\xo$ as input and yields a per-record correction $\delta_i = g(\vec{x}_{o})$.
The result is an adjusted feature that we denote as $\xadj = \xp + g(\Xother)$.

Let $D(\cdot, Z)$ denote a statistic measuring a feature's proxy capacity for demographic group $Z$.
In this paper, $D$ is the statistic of \emph{the distributional proxy test}, $\hat\kappa$ (Equation~\eqref{eq:binned}).
We note, however, that $D$ is parameterizable, and does not necessarily need to be exactly the measure of proxy capacity introduced in Section~\ref{sec:proxy-test}.

We seek the adjuster $g$ that minimizes the objective
\begin{equation}\label{eq:adjuster_objective}
    \mathcal{L}(g;\, \xp, Z)
    \;:=\; \underbrace{D\big(\xp + g(\Xother),\, Z\big)}_{\text{fairness loss}} \; +\; \underbrace{\lambda_c \cdot \lVert g(\Xother) \rVert}_{\text{complexity loss}},
\end{equation}
where the complexity weight $\lambda_c$ trades the reduction in $D$ against the size of the correction.
In $\mathcal{L}(g;\, \xp, Z)$, the adjuster $g$ is optimized conditional on $\xp$ and $Z$, which are only used in calculating its loss.
Three desiderata follow:
\begin{enumerate}
    \item[(i)] \emph{Out-of-sample generality.} $g$ should learn an adjustment that generalizes beyond the observed data.
    \item[(ii)] \emph{Signal preservation.} $\xadj$ should retain some of its task relevant signal that originally motivated the use of $\xp$. The comparison we make throughout is some downstream model $f$ trained on $(\Xother, \xadj)$ against $f$ trained on $\Xother$ alone, establishing the performance gap between \emph{removal} and the proposed framework (see Figure~\ref{fig:feature_workflow}).
    \item[(iii)] \emph{No demographic data at inference.} $Z$ is never used after training: $g$ is learned offline and applied online as a sample-dependent transformer of $\xp$, and the downstream model only sees $(\Xother, \xadj)$.
\end{enumerate}

\begin{figure}[t]
    \centering
    \includegraphics[width=0.65\textwidth]{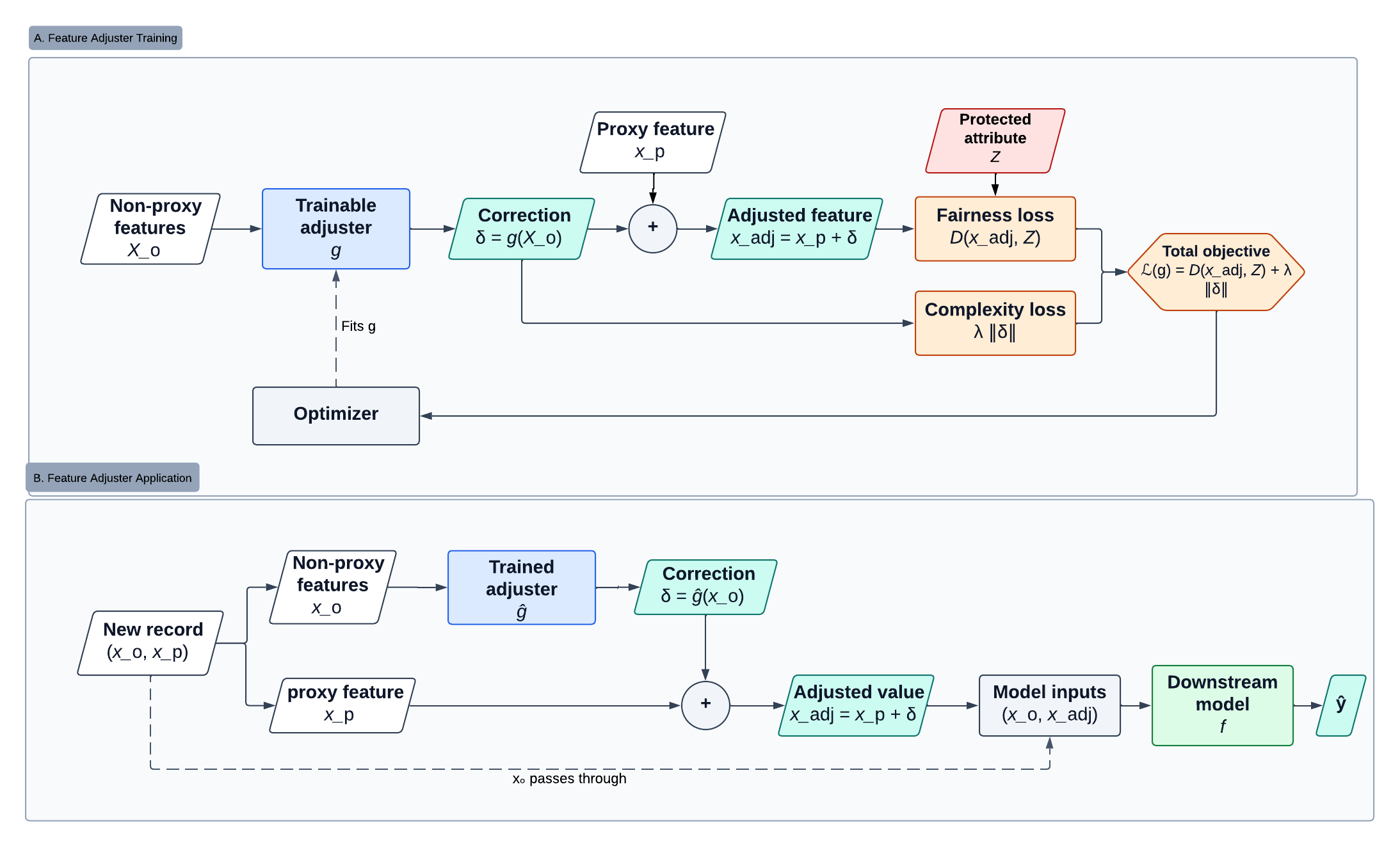}
    \Description{Two-panel diagram. Panel A, training: the non-proxy features feed a trainable adjuster g that outputs a correction delta; delta is added to the proxy feature to form the adjusted feature; a fairness loss on the adjusted feature and the protected attribute Z, plus a complexity loss on delta, form the total objective that an optimizer uses to fit g. Panel B, application: for a new record, the trained adjuster maps the non-proxy features to a correction that is added to the proxy feature, and the downstream model receives the non-proxy features and the adjusted value; Z is not used.}
    \caption{
        Feature adjuster training and application. The feature adjuster acts as a sample-dependent transformer on the feature $\xp$: we find the adjuster $g$ that optimizes the objective given by Equation~\eqref{eq:adjuster_objective} (A).
        At inference time (B), the trained adjuster takes a new record's non-proxy features $\xo$ as input and yields $\xadj$; $Z$ is not used and the model never has direct access to $\xp$.
    }
    \label{fig:adjuster_context}

\end{figure}

This workflow is visualized in Figure~\ref{fig:adjuster_context}.
The formulations of $D$, $\lVert \cdot \rVert$, and the function class of $g$ are specified in Sections~\ref{sec:soft-proxy-loss} and~\ref{sec:synthetic}.

The key intuition underpinning this framework is a tight coupling between the proxy detection and mitigation strategies.
For example, if the practitioner's proxy detection methodology $D$ is a simple linear predictor of $Z$ from $\xp$, one should train an adjuster $g$ that minimizes the performance of a linear adversary.
However, as discussed in Section~\ref{sec:proxy-test} and shown in Section~\ref{sec:exp-synthetic}, this and other aggregate measures of proxiness can be insufficient in identifying proxy capacity.

\subsection{A Differentiable Proxy Loss}\label{sec:soft-proxy-loss}

To find a $g$ that optimizes Equation~\eqref{eq:adjuster_objective}, we need an expression that is differentiable with respect to the output of $g$.
In this section we introduce a differentiable approximation of the statistic $\hat\kappa$ of the \emph{distributional proxy test}.

Two steps of the \emph{distributional proxy test} are discontinuous and block the gradient with respect to the adjuster output $\delta_i$: region membership $\ind[x_i \in R_k]$ is a step function of $x_i$, and taking the maximum over regions is not smooth.
We relax each of these two steps as described below.

\paragraph{Soft assignment.}
Let $x_i$ be the $i$-th record of $\xadj$ and $Z_i$ the $i$-th record of $Z$; region $R_k$ of Section~\ref{sec:estimating}, for $k = 1, \dots, K$, has center $\mu_k$ and width $h$.
We replace the indicator $\ind[x_i \in R_k]$ with a Gaussian-kernel softmax over regions,
\begin{equation}\label{eq:soft_assign}
    w_{ik} \;=\; \frac{\exp\!\big(-(x_i - \mu_k)^2 / 2\tau^2\big)}{\sum_{k'=1}^{K} \exp\!\big(-(x_i - \mu_{k'})^2 / 2\tau^2\big)},
    \qquad \tau = \rho\, h,
\end{equation}
so each record distributes unit mass over the regions, $\sum_k w_{ik} = 1$, with the bandwidth $\tau$ set as a multiple $\rho$ of the region width $h$.
In the limit $\rho \to 0$, the assignment concentrates on the nearest region center and recovers the strict assignment.
We then calculate the soft record count $\tilde n_k$, the soft count of group members $\tilde s_k$, and the soft rate of group membership $\tilde p_k$ of each region:
\begin{equation}\label{eq:soft_rate}
    \tilde n_k \;=\; \sum_{i=1}^{n} w_{ik}, \qquad
    \tilde s_k \;=\; \sum_{i=1}^{n} w_{ik}\, Z_i, \qquad
    \tilde p_k \;=\; \frac{\tilde s_k}{\tilde n_k}.
\end{equation}
Since each $\tilde p_k$ is a smooth function of every $x_i$, it is differentiable with respect to every $\delta_i$.

\paragraph{Smooth maximum.}
We replace the max rate of group membership over all regions with a log-sum-exp at temperature $\tau_{\max}$ \citep{boyd2004convex},
\begin{equation}\label{eq:soft_max}
    M(\xadj, Z) \;=\; \tau_{\max} \log \sum_{k=1}^{K} \exp\!\big(\frac{\tilde p_k }{\tau_{\max}}\big),
\end{equation}
which converges to the maximum as $\tau_{\max} \to 0$.
Its gradient with respect to $\tilde p_k$ is the softmax weight of region $R_k$ at temperature $\tau_{\max}$, which approaches an indicator of the region with the highest rate as $\tau_{\max}$ becomes small.

\paragraph{The objective function.}
Together, these two relaxations give the training loss.
We substitute $M$ for $D$ in Equation~\eqref{eq:adjuster_objective}, square $M$, and use the mean squared $L_2$ norm for the complexity loss:
\begin{equation}\label{eq:soft_proxy_loss}
    \mathcal{L}(g;\, \xp, Z) \;=\; M\big(\xp + g(\Xother),\, Z\big)^2 \;+\; \lambda_c \cdot \frac{1}{n} \sum_{i=1}^{n} \delta_i^2 .
\end{equation}
%

%% file: sections/04_synthetic.tex
\section{Synthetic Validation}\label{sec:synthetic}

We validate the two steps of the framework in turn on synthetic data, where the strength of association between a feature and $Z$ is known: detection (Section~\ref{sec:exp-synthetic}) and mitigation (Section~\ref{sec:synthetic-mitigation}).
Both studies calculate the \emph{distributional proxy test} over ten regions with a rate of demographic group membership $\pi = 0.2$.
For illustrative purposes, the test is calculated with the desired strength of association $B = 19$, so $T(19, 0.2) = 0.826$.

\subsection{Detecting Proxy Capacity}\label{sec:exp-synthetic}

\begin{figure}[t]
  \centering
  \begin{minipage}[t]{0.48\textwidth}
    \vspace{0pt}
    \centering
    \captionsetup{skip=1.8pt}
    \includegraphics[width=\linewidth]{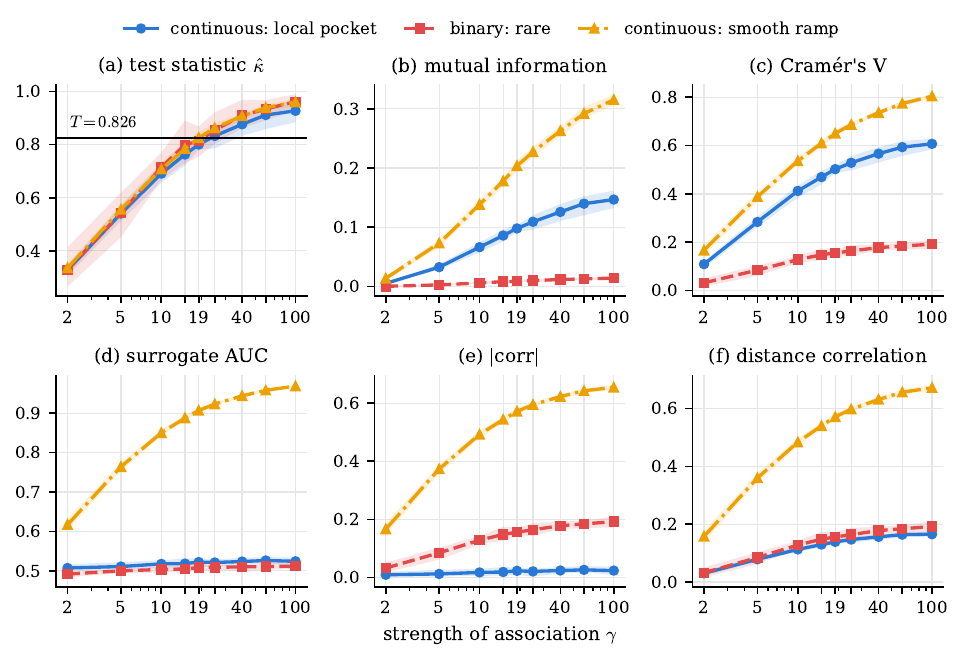}
    \Description{Six line plots against strength of association gamma for the smooth ramp, local pocket, and rare binary archetypes. In panel a, the test statistic kappa rises along the same path for all three archetypes and crosses the threshold T together. In panels b to f, mutual information, Cramer's V, surrogate AUC, absolute correlation, and distance correlation rise for the ramp but stay low for the rare and local archetypes.}
    \captionof{figure}{Response of the test statistic and the five aggregate measures of association as $\gamma$ increases.
      \textbf{(a)} $\hat\kappa$ responds the same to evidence of association, regardless of the feature's shape.
      \textbf{(b)--(f)} The aggregate measures all vary by feature shape, and all struggle to identify rare and local proxies.}
    \label{fig:response-sweep}
  \end{minipage}\hfill
  \begin{minipage}[t]{0.48\textwidth}
    \vspace{0pt}
    \centering
    \scriptsize
    \setlength{\tabcolsep}{2pt}
    \input{tables/tab_sweep}
    \par\vspace{3.9pt}
    \captionof{table}{Selected statistic means visualized in Figure~\ref{fig:response-sweep}. The other measures rise with $\gamma$ on the smooth ramp, but AUC stays near $0.5$ on the local pocket and the rare value and $|$corr$|$ stays near $0$ on the local pocket; only $\hat\kappa$ maintains a consistent scale across archetypes.}
    \label{tab:sweep}
  \end{minipage}
\end{figure}

\paragraph{Setup.} We sample five synthetic features, three continuous and two binary.
Every record $(x_i, z_i)$ is generated by first drawing the feature $X$, then drawing membership from $(Z \mid X{=}x) \sim \mathrm{Bernoulli}(p(x, \gamma))$, where the \emph{strength of association} between $X$ and $Z$ is given by $\gamma$.
For each feature, we vary $\gamma$ to study how each measure of proxiness responds in the specified feature.

The three continuous features we sample are a null (no relationship between $X$ and $Z$), a \emph{smooth ramp} (a logistic $p(x, \gamma)$ that simulates a roughly linearly correlated proxy), and a \emph{local pocket} (a discontinuous jump in the rate of group membership in $X$).
The binary features are a null and a \emph{rare} value with $\pr(X{=}1) = 0.01$.
For reference, we compute five measures of association, each of which has been previously proposed as a feature-level proxy screen: mutual information (MI) \citep{weerts2024unlawful}, Cram\'er's V \citep{cramer1946mathematical}, a surrogate AUC adapted from \citet{feldman2015certifying}, absolute Pearson correlation \citep{yeom2018hunting}, and distance correlation (dcor) \citep{szekely2007measuring}.
Further detail on the feature definitions and visualizations can be found in \ref{app:archetypes}.
We generate $n{=}10$k samples and sweep over a range of $\gamma$ for each non-null feature. We then calculate our test statistic $\hat\kappa$ and all comparison measures on these samples.

\paragraph{Reading.} Table~\ref{tab:sweep} and Figure~\ref{fig:response-sweep} show that regardless of the feature type, the test statistic $\hat\kappa$ responds directly to increasing strength of association where the other aggregate measures struggle, especially in the binary and local cases.
The proposed statistic is the only measure that is able to detect the strength of association on the same scale among all 3 feature types.
Operationalizing these measures would pose practical difficulties to practitioners tasked with designing a consistent decision framework.
The aggregate measures are built to find \emph{average} association across the population, and as such fail to identify rare or local proxy features.

\subsection{Reducing Proxy Capacity}\label{sec:synthetic-mitigation}

We next test the adjuster on synthetically generated data, where we manipulate two elements: the \emph{proxiness} or strength of relationship between $\xp$ and $Z$, and the \emph{predictiveness} or how much $\xp$ helps predict $Y$.
Based on changing the \emph{proxiness} of the feature, we can first check that adjusted features pass the \emph{distributional proxy test}, and also observe what the adjustment does to non-proxy features.
By changing the \emph{predictiveness}, we can show how much is retained through the adjusted feature vs. removal.

\paragraph{Setup.}
In this experiment, we utilize the same ``smooth ramp'' feature from Section~\ref{sec:exp-synthetic} as our experimental proxy $\xp$.
Here we also utilize $d = 20$ other, non-proxy features $\Xother$.
We create a necessary set of complex proxies by using two columns in $\Xother$, which are features jointly related to $Z$ but are not themselves classified as proxies by the \emph{distributional proxy test}.
Four additional features carry signal for $Y$, and the rest are noise.
For this experiment, we use a binary target $Y$.
We vary the \emph{predictiveness} (parameterized as $\beta_x$) and \emph{proxiness} (parameterized as $\gamma$) of $\xp$ to demonstrate the ability of the feature adjuster.

We fit two adjusters to the differentiable proxy loss of Equation~\eqref{eq:soft_proxy_loss}.\footnote{We set ten regions $K = 10$ and $\rho$ annealed from $5$ to $0.5$ in the differentiable proxy loss.}
The first is a two-hidden-layer Multi-Layer Perceptron (MLP) trained with Adam at learning rate $10^{-3}$ for 1500 steps.
The second is a gradient-boosted tree ensemble (XGB) with 400 rounds at step size $0.15$.
All results are computed on the test fold.
Using the same \emph{distributional proxy test} of Section~\ref{sec:estimating}, we compute $\hat\kappa$ over deciles of the feature and set the required strength of evidence to identify a feature as a proxy at the illustrative level $B = 19$ (Section~\ref{sec:thresholds}).

We illustrate a simple downstream model $f$ using a gradient-boosted classifier (100 trees, depth 4).
We train it using three datasets: $(\Xother, \xp)$, $(\Xother, \xadj)$, and $\Xother$ alone.
We report the test AUC of the first two relative to the third.
A value of $0$ means the feature is worth as much as removing it, and the untouched feature's value is what there is to keep.

As a control we also add white noise to the feature, $\xp + \sigma \varepsilon$ with $\varepsilon \sim \mathcal{N}(0, 1)$ and varying strength $\sigma$.
Since the noise carries no information about $Z$ or $Y$, it lowers $\hat\kappa$ only by blurring every record.

\begin{figure}[t]
  \centering
  \includegraphics[width=0.7\textwidth]{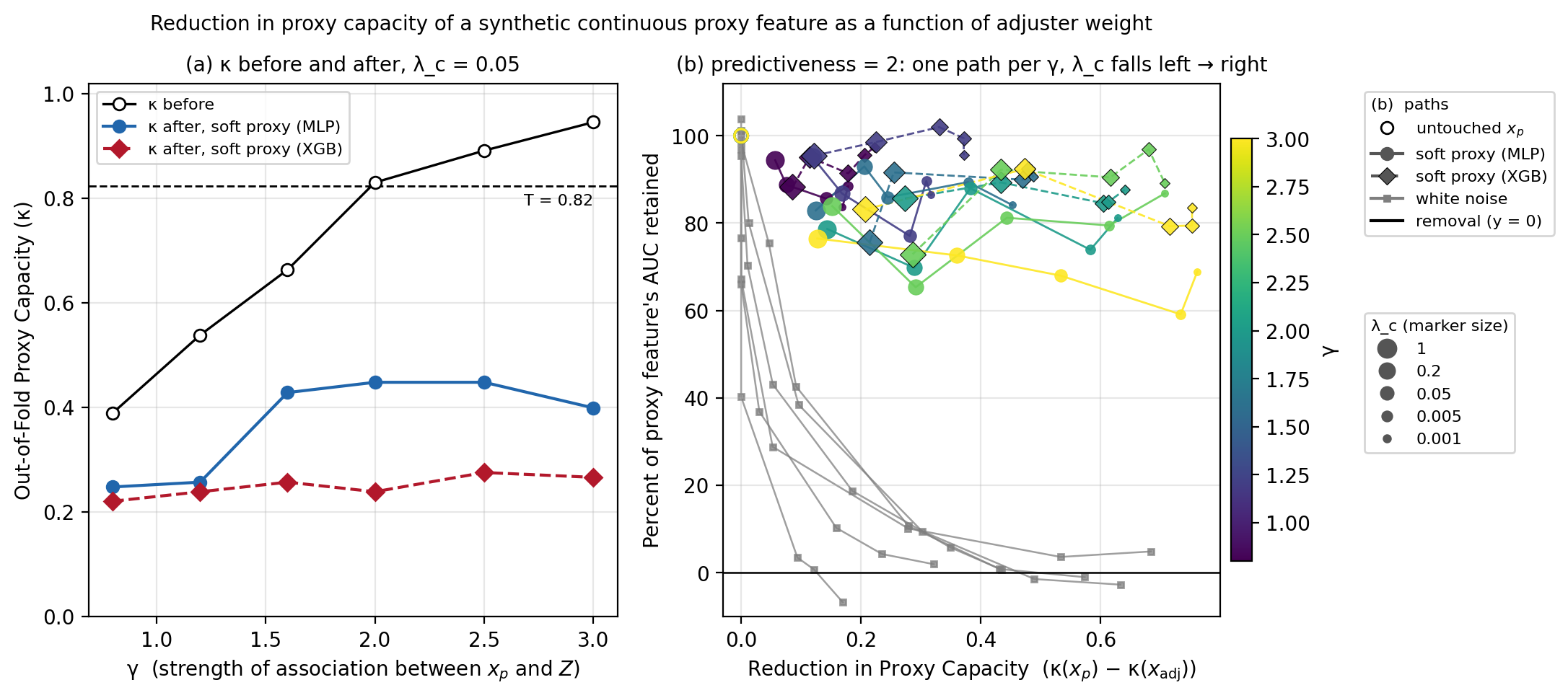}
  \Description{Two panels for the continuous synthetic feature. Panel a plots kappa before and after adjustment against gamma, with the threshold T; after adjustment every configuration is below T. Panel b plots test AUC over removal against the reduction in kappa at predictiveness 2; adjuster paths for each gamma stay well above zero as the reduction grows, while the white-noise control paths fall toward zero.}
  \caption{
    The effect of adjusting proxy features on their proxy capacity and predictiveness.
    (a) Proxy capacity $\hat\kappa$ on the test fold before and after adjustment at $\lambda_c = 0.05$.
    (b) The trade-off at high predictiveness: the horizontal axis is the reduction in proxy capacity $\hat\kappa$, and the vertical axis is \% AUC retained vs removal.
    Color is strength of association between $\xp$ and $Z$: $\gamma$.
    Each colored line is one $\gamma$ under one adjuster, solid with circles for the MLP and dashed with diamonds for XGB.
    Along a path, the complexity weight $\lambda_c$ falls from $1$ to $10^{-3}$ left to right, and marker size shrinks with it.
    Gray paths are the white-noise control increasing left to right.
  }
  \label{fig:continuous-frontier}
\end{figure}

\begin{table}[t]
  \centering
  \footnotesize
  \setlength{\tabcolsep}{4pt}
  \input{tables/tab_continuous}
  \caption{
    Continuous synthetic feature at $\lambda_c = 0.05$, medians over three seeds.
    $\hat\kappa$ is the test statistic on the test fold, and \emph{proxy} marks the configurations the test identifies as proxies at $B = 19$.
    AUC over removal is the test AUC of the downstream model with the feature, minus its test AUC with the feature removed.
    Each such column is read at one predictiveness, $\beta_x = 0.5$ or $\beta_x = 2$.
    The columns under \emph{untouched} give this quantity for the untouched feature, and the columns under each adjuster give it for the adjusted feature.
    The percentage in parentheses is the adjusted feature's AUC over removal as a share of the untouched feature's.
    $\hat\kappa$ does not depend on predictiveness and is printed once.
  }
  \label{tab:continuous}
\end{table}

\paragraph{Reading}
By construction, $\hat\kappa$ increases with $\gamma$ and crosses $T$ at $\gamma = 2.0$ (Table~\ref{tab:continuous}).
So the three largest values of $\gamma$ are proxy features under the test, and the three smallest are not.
Using an adjuster weight of $\lambda_c = 0.05$, the adjusted feature's proxy capacity $\hat\kappa$ decreased by $35\%$ to $72\%$ depending on the adjuster class and strength of association (Figure~\ref{fig:continuous-frontier}a).
Every proxy feature passes the \emph{distributional proxy test} after adjustment, and non-proxy features also reduce their proxy capacity $\hat\kappa$ after adjustment.

For proxies that had higher predictiveness, the untouched proxy feature adds $+0.042$ to $+0.051$ AUC over removal.
After adjustment, the feature keeps $68\%$ to $88\%$ of the untouched feature's gain over removal with the MLP and $79\%$ to $90\%$ with XGB (percentages in Table~\ref{tab:continuous}).

We also see that the adjuster is effective at reducing the proxy capacity of the proxy feature regardless of the strength of association between $\xp$ and $Z$ (see Figure~\ref{fig:continuous-frontier}a).
Additionally, varying the complexity weight $\lambda_c$ from $1$ to $10^{-3}$ moves each path in Figure~\ref{fig:continuous-frontier}b to the right: smaller adjuster penalties correspond to increased reductions in proxy capacity.
We also see that the adjuster achieves the same reduction in proxy capacity as adding noise, but without the degradation to downstream model performance.

%% file: tables/tab_sweep.tex
\begin{tabular}{lcccccc}
    \toprule
    archetype (strength $\gamma$) & $\hat \kappa$ & MI    & Cram\'er's V & AUC  & $|$corr$|$ & dcor \\
    \midrule
    \multicolumn{7}{l}{\emph{continuous: smooth ramp}}                                              \\
    \quad $\gamma = 2$            & 0.31          & 0.014 & 0.17         & 0.62 & 0.17       & 0.16 \\
    \quad $\gamma = 19$           & 0.81          & 0.204 & 0.65         & 0.91 & 0.57       & 0.57 \\
    \quad $\gamma = 40$           & 0.89 & 0.263 & 0.74         & 0.94 & 0.62       & 0.63 \\
    \quad $\gamma = 100$          & 0.95 & 0.315 & 0.80         & 0.97 & 0.65       & 0.67 \\
    \midrule
    \multicolumn{7}{l}{\emph{continuous: local pocket}}                                             \\
    \quad $\gamma = 2$            & 0.30          & 0.005 & 0.11         & 0.51 & 0.01       & 0.03 \\
    \quad $\gamma = 19$           & 0.78          & 0.098 & 0.50         & 0.52 & 0.02       & 0.14 \\
    \quad $\gamma = 40$           & 0.86 & 0.126 & 0.57         & 0.52 & 0.02       & 0.16 \\
    \quad $\gamma = 100$          & 0.91 & 0.147 & 0.61         & 0.52 & 0.02       & 0.17 \\
    \midrule
    \multicolumn{7}{l}{\emph{binary: rare}}                                                         \\
    \quad $\gamma = 2$            & 0.26          & 0.001 & 0.03         & 0.49 & 0.03       & 0.03 \\
    \quad $\gamma = 19$           & 0.74          & 0.009 & 0.16         & 0.51 & 0.16       & 0.16 \\
    \quad $\gamma = 40$           & 0.85 & 0.012 & 0.18         & 0.51 & 0.18       & 0.18 \\
    \quad $\gamma = 100$          & 0.92 & 0.014 & 0.19         & 0.51 & 0.19       & 0.19 \\
    \bottomrule
\end{tabular}

%% file: tables/tab_continuous.tex

\begin{tabular}{r cc rr crr crr}
    \toprule
             & \multicolumn{4}{c}{untouched} & \multicolumn{3}{c}{soft proxy (MLP)} & \multicolumn{3}{c}{soft proxy (XGB)}                                                                                                                                                                    \\
    \cmidrule(lr){2-5}\cmidrule(lr){6-8}\cmidrule(lr){9-11}
             &                               &                                      & \multicolumn{2}{c}{AUC over removal} &               & \multicolumn{2}{c}{AUC over removal} &                 & \multicolumn{2}{c}{AUC over removal}                                                    \\
    \cmidrule(lr){4-5}\cmidrule(lr){7-8}\cmidrule(lr){10-11}
    $\gamma$ & $\hat \kappa$                 & proxy                                & $\beta_x{=}0.5$                      & $\beta_x{=}2$ & $\hat \kappa$                        & $\beta_x{=}0.5$ & $\beta_x{=}2$                        & $\hat \kappa$ & $\beta_x{=}0.5$ & $\beta_x{=}2$  \\
    \midrule
    0.8      & 0.39                          & $\cdot$                              & +0.015                               & +0.065        & 0.25                                 & +0.013 (88\%)   & +0.056 (86\%)                        & 0.22          & +0.007 (45\%)   & +0.060 (91\%)  \\
    1.2      & 0.54                          & $\cdot$                              & +0.009                               & +0.059        & 0.26                                 & +0.002 (21\%)   & +0.046 (77\%)                        & 0.24          & +0.006 (61\%)   & +0.060 (102\%) \\
    1.6      & 0.66                          & $\cdot$                              & +0.013                               & +0.058        & 0.43                                 & +0.006 (45\%)   & +0.050 (86\%)                        & 0.26          & +0.009 (71\%)   & +0.052 (90\%)  \\
    2        & 0.83                          & \checkmark                           & +0.012                               & +0.051        & 0.45                                 & +0.012 (104\%)  & +0.045 (88\%)                        & 0.24          & +0.008 (67\%)   & +0.043 (85\%)  \\
    2.5      & 0.89                          & \checkmark                           & +0.003                               & +0.042        & 0.45                                 & +0.004 (137\%)  & +0.034 (81\%)                        & 0.28          & +0.001 (39\%)   & +0.038 (90\%)  \\
    3        & 0.95                          & \checkmark                           & +0.011                               & +0.044        & 0.40                                 & +0.008 (75\%)   & +0.030 (68\%)                        & 0.27          & +0.006 (57\%)   & +0.035 (79\%)  \\
    \bottomrule
\end{tabular}

%% file: sections/05_real_world.tex
\section{Real-World Illustration}\label{sec:real-world}

In this section we aim to illustrate how to incorporate this framework in two practical settings.
First, in Section~\ref{sec:real-world-audit} we illustrate how to operationalize proxy feature detection using the \emph{distributional proxy test} on the UCI dataset.
Second, in Section~\ref{sec:real-world-cell} we illustrate how operationalizing the feature adjuster reduces proxy capacity while still improving model performance over removal on the Folktables ACS dataset.

We chose these datasets for consistency with previous literature and purely for illustrative purposes: they provide a setting that mirrors what ML practitioners commonly face when applying ML models in regulated domains such as lending.

\subsection{Detecting Proxy Capacity in the UCI Dataset}\label{sec:real-world-audit}

\paragraph{Setup.}
\citet{weerts2024unlawful} (Section~4.1.1) score the features of the UCI Adult dataset \citep{uci_adult} with mutual information and report that the ``Relationship'' feature is a proxy for ``Sex'' and ``Native-country'' as a proxy for ``Race''.
We apply the \emph{distributional proxy test} to the same data.
To classify features as ``High'' proxy risk, one could set the strength of evidence required $B$ and calculate thresholds per demographic group according to Equation~\eqref{eq:thresh}.

\paragraph{Reading.}
Using the \emph{distributional proxy test}, we found that ``Relationship $=$ Husband'' is a proxy for ``male'', and ``Relationship $=$ Wife'' is a proxy for ``female''.
This overcomes one of the problems with using MI to measure association between a feature and a protected characteristic with two groups (i.e. male vs. female): it is bi-directional, so the MI is the same for both demographic groups.
Additionally, setting a strict threshold on MI alone fails to identify which values of ``native-country'' proxy for which ``Races''.
For example, we found that ``native-country $=$ United-States'' has higher MI with ``Race $=$ Asian-Pacific Islander'' than all the other values that were flagged by the \emph{distributional proxy test} such as ``China'', ``India'', and ``Vietnam.''
This clearly highlights the finding from Section~\ref{sec:exp-synthetic}, where aggregate measures of association are not sensitive to rare binary features.
We provide a reference of the \emph{distributional proxy test} results in Table~\ref{tab:adult-values} in \ref{app:audit}.

\subsection{Reducing Proxy Capacity in the Folktables Dataset}\label{sec:real-world-cell}

\paragraph{Setup.}
For this experiment, we used the Folktables ACS Income for California 2018 \citep{ding2021retiring}, which is commonly used to build example ML models that target whether an individual's income exceeds \$50{,}000.
We identified the feature \texttt{SCHL}, educational attainment on 24 ordered levels from no schooling to a doctorate, as a proxy feature for the demographic group \texttt{RAC1P} $= 8$, or ``Race $=$ `some other race alone' '' using the \emph{distributional proxy test}.
Specifically, we found that the lowest educational attainment revealed membership in this demographic group.
For illustrative purposes, we set the strength of evidence at $B = 3$, the conventional level for evidence worth reporting \citep{kass1995bayes}.
Among records without a high-school diploma, the lower confidence bound on the group's share is $0.29$ against a threshold of $T(3, 0.117) = 0.28$, so the test identifies the feature as a proxy (implied Bayes factor $3.2$).
Removing it lowers the classifier's test AUC from $0.896$ to $0.887$.
We followed the same procedure established in Section~\ref{sec:synthetic-mitigation}: we trained both MLP and XGBoost feature adjusters using the other non-proxy features with varying complexity weights $\lambda_c$, and then assessed the proxy capacity $\hat\kappa$ out of fold before training a classifier using $\xadj$.
We also provided a control by adding white noise to $\xp$ for reference.

\begin{figure}[t]
    \centering
    \begin{minipage}[t]{0.48\textwidth}
        \vspace{0pt}
        \centering
        \includegraphics[width=\linewidth]{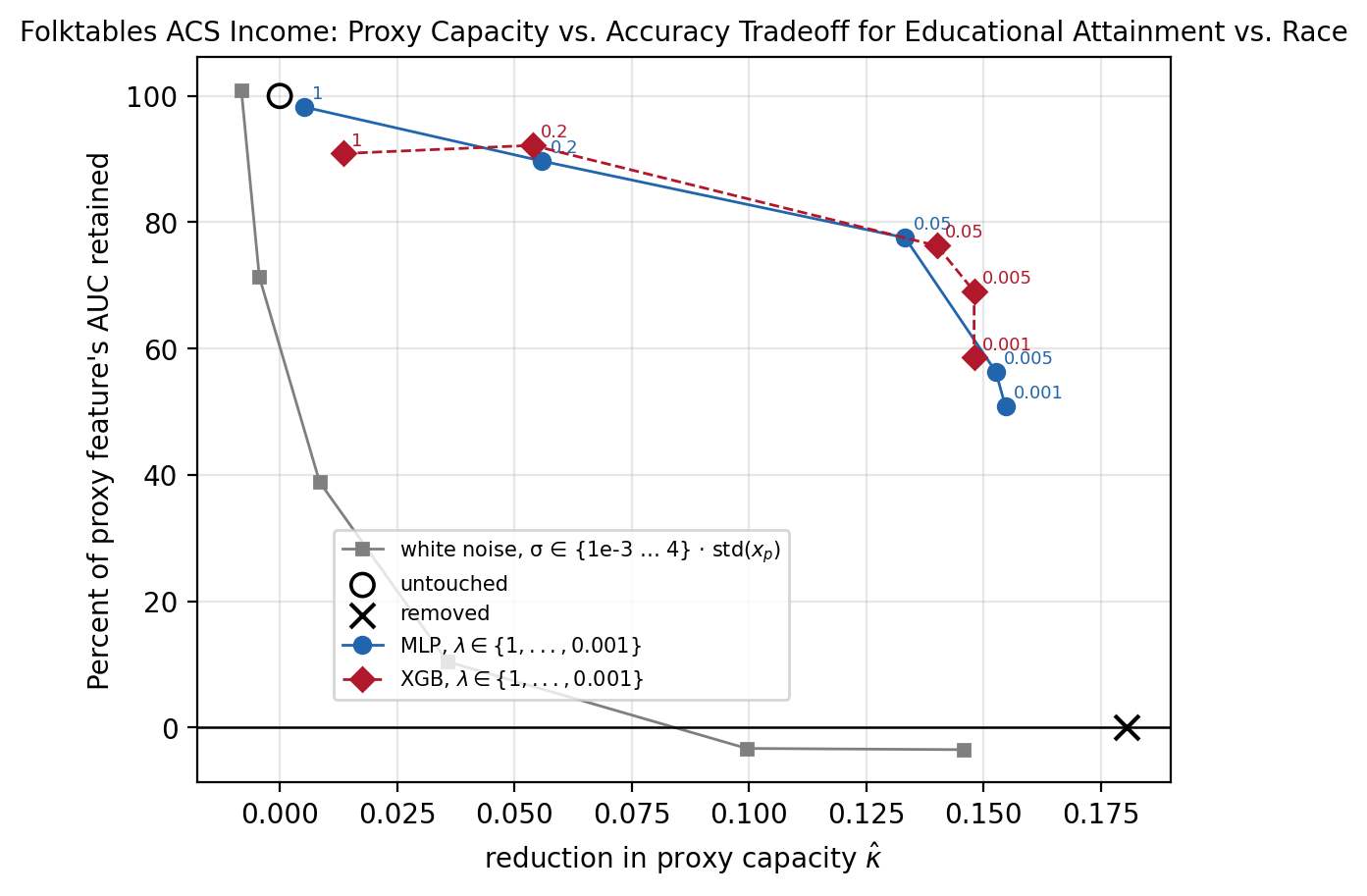}
        \Description{Scatter plot for educational attainment in Folktables: reduction in kappa on the horizontal axis and test AUC over removal on the vertical axis. MLP and XGB adjuster points, labeled by lambda, lie above the white-noise points at every reduction, between the untouched feature and removal.}
        \captionof{figure}{
            Educational attainment (\texttt{SCHL}) against \texttt{RAC1P} $=$ ``some other race'' in Folktables ACS Income.
            The horizontal axis is the reduction in proxy capacity $\hat\kappa$ on the test fold, and the vertical axis is test AUC over removal.
            Blue circles are the MLP adjuster and red diamonds are the XGB adjuster.
            Each marker is labeled with its complexity weight $\lambda_c$, which falls from $1$ to $10^{-3}$ left to right.
        }
        \label{fig:schl-frontier}
    \end{minipage}\hfill
    \begin{minipage}[t]{0.50\textwidth}
        \vspace{0pt}
        \centering
        \scriptsize
        \setlength{\tabcolsep}{3pt}
        \input{tables/tab_schl_compact}
        \captionof{table}{
            Each row is one version of the feature: untouched, with white noise added, adjusted under each adjuster, or removed.
            The column $\hat\kappa$ after is the test statistic on that version, and implied BF is its implied Bayes factor.
            AUC $-$ removal is the downstream model's test AUC minus the same seed's removal AUC.
            The adjusters decrease the proxy capacity $\hat\kappa$ while keeping most of the improvement to AUC.
        }
        \label{tab:schl}
    \end{minipage}
\end{figure}

\paragraph{Reading.}
For both adjuster classes, we can see that the complexity weight $\lambda_c$  traces a path from the untouched feature toward removal (Figure~\ref{fig:schl-frontier}): smaller complexity weights lower proxy capacity more while degrading predictive performance.
At $\lambda_c = 0.2$, $0.05$, and $10^{-3}$ with the MLP, the adjusted feature keeps between $51\%$ and $89\%$ of the untouched feature's AUC over removal (Table~\ref{tab:schl}).
We do not note any significant differences between adjuster classes: the impact to proxy capacity $\hat\kappa$ and downstream AUC is not meaningfully different between the XGB and MLP at every $\lambda_c$.

The noise control shows that the reduction in proxy capacity yields the same degradation in accuracy as feature removal.
Every point on the adjusters' paths Pareto dominates the white noise curve.

%% file: tables/tab_schl_compact.tex

\begin{tabular}{l rrr}
    \toprule
    feature & $\hat \kappa$ after & implied BF & AUC $-$ removal \\
    \midrule
    untouched & 0.294 & 3.16 & +0.0102 \\
    \addlinespace
    noise, $\sigma = 0.25$ & 0.299 & 3.23 & +0.0072 \\
    noise, $\sigma = 0.5$  & 0.285 & 3.03 & +0.0039 \\
    noise, $\sigma = 1$    & 0.256 & 2.61 & +0.0011 \\
    noise, $\sigma = 2$    & 0.195 & 1.83 & -0.0003 \\
    noise, $\sigma = 4$    & 0.149 & 1.32 & -0.0004 \\
    \addlinespace
    MLP, $\lambda_c = 1$      & 0.291 & 3.11 & +0.0100 \\
    MLP, $\lambda_c = 0.2$    & 0.240 & 2.40 & +0.0091 \\
    MLP, $\lambda_c = 0.05$   & 0.161 & 1.45 & +0.0079 \\
    MLP, $\lambda_c = 0.001$  & 0.140 & 1.23 & +0.0052 \\
    \addlinespace
    XGB, $\lambda_c = 1$      & 0.277 & 2.91 & +0.0092 \\
    XGB, $\lambda_c = 0.2$    & 0.241 & 2.40 & +0.0094 \\
    XGB, $\lambda_c = 0.05$   & 0.154 & 1.38 & +0.0077 \\
    XGB, $\lambda_c = 0.001$  & 0.144 & 1.28 & +0.0059 \\
    \addlinespace
    removed & 0.114 & 0.97 & +0.0000 \\
    \bottomrule
\end{tabular}

%% file: sections/06_discussion.tex
\section{Discussion}\label{sec:discussion}

Here we highlight possible directions of future research that extend this framework in other domains and the limitations of this paper.

\paragraph{Application to Language Model Systems}
Systems built around language models (LMs) are currently being applied in a variety of domains, and emerging literature is exploring applications in regulated domains, such as lending \citep{bowen2024measuring, cook2025social}.
Prior work has shown that LMs can perpetuate biases based on both overt and covert demographic group membership in hypothetical situations: for example,
\citet{hofmann2024dialect} find that LMs assign systematically worse outcomes to speakers of African American English, based on their dialect alone.
Further, \citet{kondapally2026sidebyside} find that covert, dialect-based bias increases when texts of different dialects are compared side by side.
\citet{anthis2025impossibility} argue that rigorous fairness definitions, including fairness through unawareness, cannot be satisfied by general-purpose LMs, in part because LMs can infer demographic group membership from its inputs.
This suggests that when LMs use inputs that reveal group membership, their use in regulated domains may raise disparate treatment concerns that mirror those raised by the use of proxy features in traditional machine learning models \citep{weerts2024unlawful}.

\begin{figure}[t]
    \centering
    \includegraphics[width=0.7\textwidth]{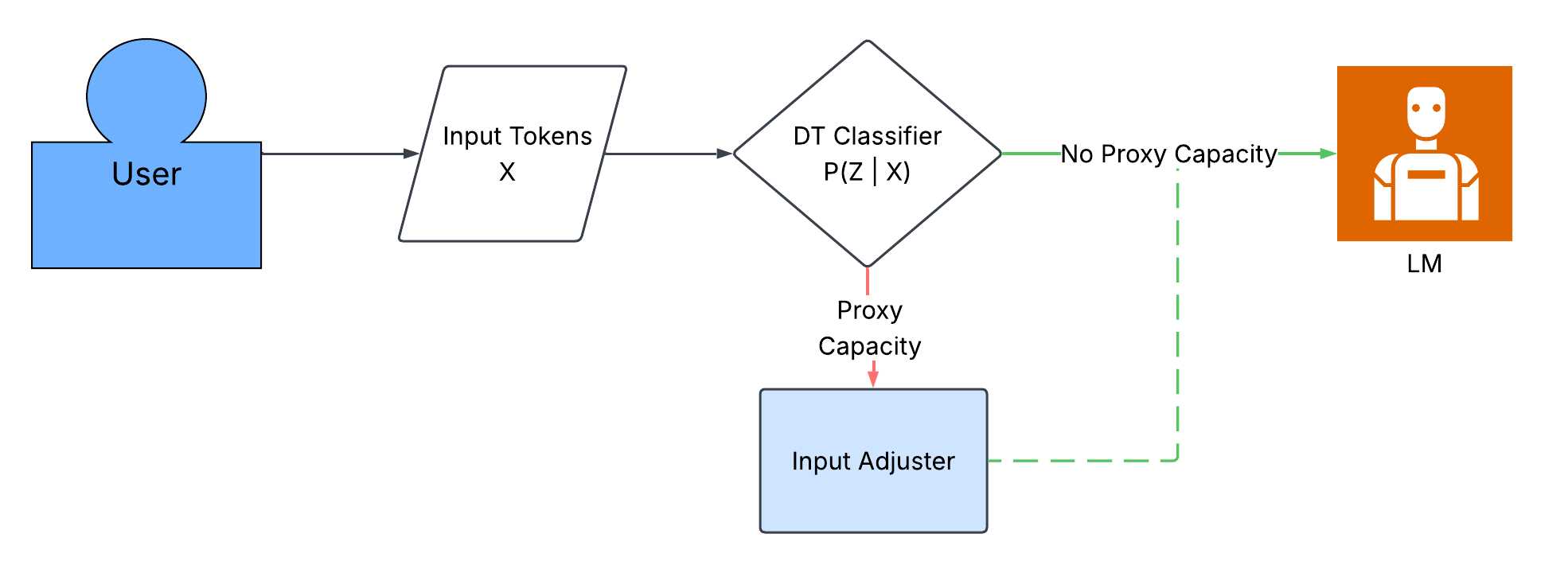}
    \Description{Flow diagram. A user produces input tokens X. A classifier of the probability of group membership Z given X tests the input. If the input has no proxy capacity, it goes directly to the language model. If it has proxy capacity, it goes to an input adjuster, whose output goes to the language model.}
    \caption{
        The proposed framework applied online to an LM system.
        The text of the user is the input tokens $X$.
        The DT (disparate treatment) classifier estimates $\pr(Z \mid X)$, which is the analogue of the \emph{distributional proxy test} in this situation.
        An input with proxy capacity goes to the input adjuster, which aims to reduce that capacity while preserving task relevant content, and the adjusted input (dashed) passes to the LM.
    }
    \label{fig:lm-discussion}
\end{figure}

In this paper, we proposed this framework as an offline ablation for proxy features; however,
the same logic can be applied online to the inputs of LMs.
For example, one could use this approach to identify disparate treatment caused by covert discrimination based on the dialect of an input.
Adapting the definition for proxy capacity established in Equation~\eqref{eq:z_given_x}, a practitioner can build a classifier assessing $\pr (Z \mid X)$ where $X$ now is the set of input tokens to the LM system (Figure~\ref{fig:lm-discussion}).
If the test flags an input, an adjustment can be made that aims to preserve the task relevant signal (i.e. semantic meaning) while reducing the proxy capacity, for example by removing overt group references or attempting to transpose the input into ``standard'' dialect.
Of course, attempting to do so without deteriorating performance is not without its own challenges: rewriting queries into Standard American English has not been extensively evaluated for fidelity \citep{klisura2025multiagent}.
Further research is necessary to explore possible proxy mitigation strategies for LM inputs.

There already exists a practical precedent for this approach in safety classifiers: classifiers are used to screen LM traffic and escalate flagged exchanges that represent safety risks such as jailbreaks and other harms \citep{cunningham2026constitutional}.

\paragraph{Regulatory compliance.}
We do not claim that an adjusted feature is broadly compliant with all anti-discrimination legislation.
The proposed methodology mirrors other established offline uses of demographic information (using $Z$ offline during model development); however, it has not been definitively established whether these approaches are permissible legally \citep{barocas2016big, weerts2024unlawful}.
Additionally, the degree to which a feature is related to demographic group membership is not clearly defined, so practitioners are left to use their best judgement when specifying their use of the \emph{distributional proxy test}.
Further legal scholarship is necessary to continue to clarify the regulatory requirements surrounding this approach.

\paragraph{Simple vs. complex proxy capacity.}
As discussed above, the proposed approach addresses simple proxy capacity as proposed by \citet{weerts2024unlawful}: the capacity of one feature to reveal $Z$.
However, a set of several features can each individually fall below $T$ and still reveal $Z$ jointly.
For example, in the case of lending, when the distribution of the creditworthiness differs systematically by demographic group, then it is likely to be the case that complex proxies are in fact critical to predict that target \citep{corbett2023measure}.
This is generally why it has been argued to consider both disparate impact and disparate treatment when constructing policies to assess the fairness of ML based systems \citep{weerts2024unlawful, black2026fairlending}.
These policies aim to mitigate the risks posed by complex proxies through their detection and mitigation of disparate impact \citep{gillis2024operationalizing}.

%% file: sections/08_conclusion.tex
\section{Conclusion}\label{sec:conclusion}

This paper addresses some of the common problems practitioners face while attempting to apply disparate treatment regulations to ML based systems.
Practitioners surveyed in \citet{black2026fairlending} echoed many of the directions for future research established in \citet{weerts2024unlawful}, specifically how to operationalize the detection of proxy features.

To address those concerns, we proposed the \emph{distributional proxy test} to identify if certain values of some feature $X$ reveal membership in demographic group $Z$ by assessing  $\pr(Z{=}1 \mid X) \geq T$.
We illustrate how to select the threshold $T$ to ensure that the same strength of evidence is required regardless of the relationship between $X$ and $Z$ or how rare the demographic group of interest is.
This approach is easily parameterizable by the practitioner to reflect their organization's position and domain expertise, and easily can be used to classify features as Low, Medium, or High risk.
Additionally, we propose a methodology that gives practitioners a middle path between exclusion or unmodified inclusion by searching for a per-feature adjuster $g$ that can learn an adjustment to the proxy feature that reduces its proxy capacity.
This approach remains consistent with other previously accepted approaches such as adversarial debiasing that mitigate fairness disparities without the use of demographic information at inference time.
We demonstrate the effectiveness of this approach by identifying proxy features that would otherwise be missed by aggregate measures of association.
Subsequently, we show how the predictive signal added by these proxy features can be retained after adjustment that mitigates their proxy capacity.

With increasing adoption of AI systems in wider domains, we believe that it is imperative to develop quantitative frameworks to detect when inputs reveal membership in protected demographic groups to prevent algorithmic discrimination.

%% file: sections/99_endmatter.tex
\section*{Generative AI Usage Statement}
The author used Claude Opus 5.5 (Anthropic), through the Claude Code tool, during manuscript preparation to identify relevant literature, and condense text to the page limit.
Claude Opus 5.5 was also used to format the document and bibliography and to cross reference the manuscript against the submission guidelines.
Claude Code was also used to assist with implementing parts of the experiment code.
The author reviewed and human-edited all generated text, verified every quotation and claim, and takes full responsibility for the content.

\section*{Acknowledgements}
The author would like to acknowledge John Lewis and Leander Eberhard for their contributions and insight in developing this framework.
Praveen Venkatesh, Sejal Aggarwal, and Lahin Lalani all provided valuable feedback while drafting this manuscript.

\section*{Competing Interests}
The author is employed by Upstart Network, Inc., which operates an AI-based lending marketplace and provides technology and related services to lending partners. Upstart funded this research and has a commercial interest in the methods proposed in this paper, including pending patent applications covering certain technology and methods described here.

\section*{Positionality Statement}
The author is an ML practitioner at a US consumer lending platform that develops AI systems which  are subject to US fair lending regulations.
This position provided the practical motivation to develop this methodology: our experience necessitated a consistent, quantitative, and scalable method to identify proxy features.
However, this position also limits this work: the proposed methodology builds upon a US-centric view of fair lending from a lender's perspective.
We acknowledge that this stands in contrast to loan applicants' perspective.
Loan applicants and the communities most affected by proxy discrimination have less power over these systems than lenders do, and their perspectives are not directly represented here.
We also acknowledge that this paper reflects the limited perspective of the author, so throughout this work, we aimed to broaden this perspective by framing this work in terms of previously published surveys and interviews like \citet{black2026fairlending}.

\section*{Ethical Considerations Statement}
This work does not describe experiments with human subjects, users, or deployed systems.
All experiments use synthetic data or public datasets (Adult, Folktables ACS Income), and the identities of individuals in these datasets are unknown.
The author recognizes the shortcomings of the Adult dataset \citep{ding2021retiring}; the paper uses it because \citet{weerts2024unlawful} use it to illustrate proxy capacity.
Demographic data or estimates of demographic data are only used offline to test for proxiness and to train the adjuster model; the downstream model never receives them.

\section*{Intellectual Property Notice}
Certain technology and methods described in this paper are subject to pending patent protection by Upstart Network, Inc.

%% file: sections/A_uncertain_z.tex
\section{Uncertain Demographic Estimates}\label{app:uncertain-z}

Often in regulated domains such as lending, demographic labels are not observed, so practitioners often use probabilistic estimates: $\hat Z_i \in [0,1]$ in place of the unobserved true labels $Z_i \in \{0,1\}$ \citep{cfpb2014bisg, elliott2009bisg}.
This introduces bias into estimating $\hat \kappa$, so we therefore need to account for this through adjusting the threshold $T$ \citep{chen2019fairness, kallus2022assessing, elzayn2023estimating}.

For a region $R$, the test estimates the rate of group membership by $\hat p_R = \frac{s_R}{n_R}$ (Section~\ref{sec:estimating}), so when labels are probabilistic $s_R$ is itself an expected count.
The true object of interest is $p_R = \pr(Z{=}1 \mid X \in R)$.
Thus the bias induced from using probabilistic estimates is
\begin{equation}\label{eq:bias}
  \Delta_R = E[\hat p_R \mid p_R] - p_R .
\end{equation}

Consider a demographic category with two groups $1$ and $0$.
Let the mean of the demographic estimator per group be $\mu_1 = E[\hat Z \mid Z{=}1]$ and $\mu_0 = E[\hat Z \mid Z{=}0]$ respectively, so a perfect estimator has $\mu_1 = 1$ and $\mu_0 = 0$.
Since any imperfect estimator assigns non-zero positive mass to non-members, true members have a negative residual $\hat Z_i - Z_i \le 0$, while true non-members have a positive residual $\hat Z_i - Z_i \ge 0$.

\begin{assumption}[Mean-nondifferential estimation]\label{asm:nondiff}
  For every tested region $R$ and each $z \in \{0,1\}$, the estimator's group-conditional mean does not vary with the tested feature: $E[\hat Z \mid Z{=}z,\, X \in R] = E[\hat Z \mid Z{=}z] = \mu_z$.
  Equivalently, the estimation error $\hat Z - Z$ is mean-independent of $X$ given the true group $Z$.
\end{assumption}

For a region with true group-membership rate $p_R$, under Assumption~\ref{asm:nondiff} the expected value of the test's estimate is
\begin{equation}
  E[\hat p_R \mid p_R] = p_R\,\mu_1 + (1-p_R)\,\mu_0.
\end{equation}
Therefore, Equation~\eqref{eq:bias} becomes
\begin{equation}
  \Delta_R = p_R\,\mu_1 + (1-p_R)\,\mu_0 - p_R = -p_R(1-\mu_1) + (1-p_R)\,\mu_0.
\end{equation}
The first term is the downward contribution from true group members whose estimated scores are below one; the second term is the upward contribution from non-members whose estimated scores are above zero.

The bias changes sign at $p^\star = \mu_0/(1 - \mu_1 + \mu_0)$.
If the demographic estimator is calibrated to the group mean, $E[\hat Z] = \pr(Z{=}1) = \pi$, this break-even point equals the base rate: $p^\star = \pi$.
Since the threshold in the \emph{distributional proxy test} is defined by a Bayes factor $B > 1$, we have $T(B, \pi) > \pi$.
So using an imperfect demographic estimator guarantees that every region that truly satisfies the test, $p_R \ge T(B, \pi)$, has strictly negative bias: $E[\hat p_R \mid p_R] - p_R < 0$.

As an illustrative example, suppose the estimator assigns $\hat Z = 0.9$ to all true group members and $\hat Z = 0.1$ to true non-members where $\pi = 0.5$.
Consider now a region whose true group rate is $p_R = 0.95$; this region has expected estimated rate of group membership of
\begin{equation}
  p_R\,\mu_1 + (1-p_R)\,\mu_0 = 0.95 \times 0.9 + 0.05 \times 0.1 = 0.86.
\end{equation}
The true region posterior is $0.95$, but the observed posterior is only $0.86$, so without a buffer, the test would fail to identify this feature at $B = 19$.

There are many potential ways to account for this, that the practitioner can choose.
One such example would be to lower the threshold to the value a true proxy is expected to achieve after introducing uncertainty.
A region whose true rate is exactly the threshold $T(B, \pi)$ has a lower expected estimated rate of
\begin{equation}
  T_{\mathrm{mean}}
  \;=\; T(B, \pi)\,\mu_1 + \left(1 - T(B, \pi)\right)\mu_0
  \;<\; T(B, \pi).
\end{equation}
We therefore compare the point estimate $\hat\kappa$ with the buffered threshold $T_{\mathrm{buf}} = T_{\mathrm{mean}}$ in place of $T(B, \pi)$.
The equation for $T_{\mathrm{mean}}$ holds under Assumption~\ref{asm:nondiff} and requires the means $\mu_1$ and $\mu_0$ of the demographic estimator, which practitioners can measure on a sample with known labels.
For a statistic that includes a lower confidence bound, \ref{app:estimate-uncertainty} describes how to buffer that bound.

Also note, better demographic estimators need smaller adjustments since $\mu_1$ is closer to one and $\mu_0$ is closer to zero, so the gap between $T_{\mathrm{mean}}$ and $T(B, \pi)$ is smaller.
Critical in Assumption~\ref{asm:nondiff}, if the candidate feature overlaps the estimator's inputs, for example when testing ZIP code while $\hat Z$ is calculated from geography, as in BISG, the estimation fails \citep{elliott2009bisg}.

%% file: sections/A2_estimate_uncertainty.tex
\section{Uncertainty in the Estimate}\label{app:estimate-uncertainty}

Because $\hat\kappa$ is a maximum over regions (Equation~\eqref{eq:binned}), sampling noise biases it upward, most of all in regions with few records.
We recommend that practitioners consider a credible interval over the quantity of interest, instead of the point estimate.
One possible way of calculating that is described below.

We can place a Jeffreys prior, $\mathrm{Beta}(\tfrac12,\tfrac12)$, on each region's rate of group membership and take the lower bound of its posterior \citep{jeffreys1961theory} to create such an interval,
\begin{equation}\label{eq:jeffreys}
  \lcb_R \;=\; F^{-1}_{\mathrm{Beta}\left(s_R + \frac12,\; n_R - s_R + \frac12\right)}(\alpha),
  \qquad \alpha = 0.05,
  \qquad
  \lcbk \;=\; \max_{R :\, n_R \ge n_{\min}} \lcb_R ,
\end{equation}
where $F^{-1}_{\mathrm{Beta}(a,b)}$ is the inverse CDF of a Beta distribution.
The test then identifies a feature as a proxy when $\lcbk \ge T$.
When demographic labels are observed, $\lcb_R$ is the standard one-sided Jeffreys interval, whose coverage is well studied \citep{brown2001interval}.

%% file: sections/B_synthetic_archetypes.tex
\section{Synthetic Archetypes and Aggregate Measures}\label{app:archetypes}

This appendix defines the synthetic features and the comparison measures of Section~\ref{sec:exp-synthetic}.
Every record $(x_i, z_i)$ is generated by first drawing the feature $X$, then drawing membership from $(Z \mid X{=}x) \sim \mathrm{Bernoulli}(p(x, \gamma))$, with $\pi = 0.2$.

\begin{figure}[t]
  \centering
  \includegraphics[width=\textwidth]{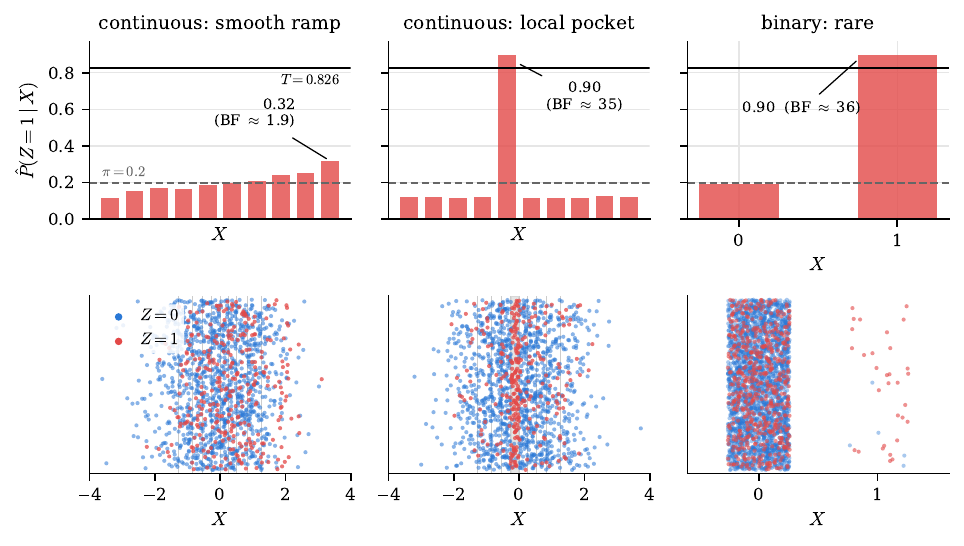}
  \Description{Three columns for the smooth ramp, local pocket, and rare binary archetypes. The top row shows the per-region rate of group membership against the base rate 0.2 and the threshold 0.826; the bottom row shows jittered scatter plots of group membership against X.}
  \caption{The three non-null synthetic archetypes. Top row: the per-region group membership rate, against the base rate $\pi{=}0.2$ (dashed) and the threshold $T{=}0.826$.
    Bottom row: a scatter plot showing group membership vs. $X$ (vertical positions are random jitter).
  }
  \label{fig:archetypes}

\end{figure}

The continuous archetypes draw $X \sim \mathcal{N}(0,1)$ and set $p(x, \gamma)$ as follows:
\begin{itemize}
  \item \emph{null}: $p(x, \gamma) = \pi = 0.2$. Group membership has no dependence on $X$.
  \item \emph{smooth ramp}: $p(x, \gamma) = 1/(1 + e^{-(a + b x)})$, a logistic ramp whose slope $b$ is set so that the top decile of $X$ has rate of group membership $T(\gamma, \pi)$, with the intercept $a$ recalibrated so the marginal stays at $\pi = 0.2$. This simulates a roughly linearly correlated proxy.
  \item \emph{local pocket}: a discontinuous jump on the fifth decile of $X$, where the strength of association is set by $\gamma$,
    \[
      p(x, \gamma) =
      \begin{cases}
        T(\gamma, \pi) & \text{if } -0.253 \le x < 0,\\[2pt]
        \bigl(\pi - 0.1\, T(\gamma, \pi)\bigr)/0.9 & \text{otherwise},
      \end{cases}
    \]
    which holds the marginal at exactly $\pi = 0.2$ for every $\gamma$.
    The pocket is centered, so $E[X \mid Z]$ stays roughly constant.
\end{itemize}
The binary archetypes draw $X \in \{0, 1\}$ and set $p(x, \gamma)$ according to:
\begin{itemize}
  \item \emph{null}: $X \sim \mathrm{Bernoulli}(0.5)$ and $p(x, \gamma) = \pi = 0.2$. Group membership has no dependence on $X$.
  \item \emph{rare}: we set this similarly to the \emph{local pocket}, now drawing $X \sim \mathrm{Bernoulli}(0.01)$, where the strength of association is again set by $\gamma$,
    \[
      p(x, \gamma) =
      \begin{cases}
        T(\gamma, \pi) & \text{if } x = 1,\\[2pt]
        \bigl(\pi - 0.01\, T(\gamma, \pi)\bigr)/0.99 & \text{if } x = 0,
      \end{cases}
    \]
    which holds the marginal at exactly $\pi=0.2$ for every $\gamma$.
\end{itemize}
See Figure~\ref{fig:archetypes} for a visualization of the three non-null archetypes.

For each feature, we compute five additional measures of association for reference, each of which has been previously proposed as a feature-level proxy screen:
\begin{itemize}
  \item Mutual Information (MI) \citep{weerts2024unlawful};
  \item Cram\'er's V \citep{cramer1946mathematical};
  \item Surrogate AUC, adapted from \citet{feldman2015certifying}: each row is scored by:
    \[
      s(x) =
      \begin{cases}
        x & \text{if $X$ is continuous},\\[2pt]
        \hat{\pr}(Z{=}1 \mid X{=}x) & \text{if $X$ is binary (out of sample)}.
      \end{cases}
    \]
    This gives us how well the feature $X$ ranks group members above non-members.\footnote{Since AUC depends only on how a score orders the rows, this gives us the probability that a randomly drawn group member is ordered above a randomly drawn non-member \citep{pepe2003statistical}.  For continuous features we take $\max(\mathrm{AUC}, 1{-}\mathrm{AUC})$ so the direction does not matter.}
  \item Absolute Pearson Correlation \citep{yeom2018hunting};
  \item Distance Correlation (dcor) \citep{szekely2007measuring}.
\end{itemize}

%% file: sections/D_real_world_audit.tex
\section{UCI Adult Results}\label{app:audit}

This appendix gives the UCI Adult results of Section~\ref{sec:real-world-audit} in full.

\paragraph{UCI Adult.}
We test every group against the twelve other features of the dataset, six categorical and six numeric, and every value of a categorical feature is one candidate.
Table~\ref{tab:adult-values} reports the flagged values for every group.
For comparison, it reports the aggregate measures of Section~\ref{sec:exp-synthetic}, which are defined in \ref{app:archetypes}.
For a binary candidate, Cram\'er's V, $|$corr$|$, and dcor coincide exactly, so one column reports all three.
Two groups have no flagged value, for opposite reasons.
For White, $T = 0.991$ and no value raises the odds of membership by a factor of $19$.
For Amer-Indian, $T = 0.156$ and no eligible value concentrates the group even that far.
Two further values of native-country, Cambodia ($28$ records) and Laos ($23$ records), both for Asian-Pacific Islander, have a lower bound above $T$ but hold fewer than $30$ records, so we did not include these results.

\begin{table}[t]
  \centering
  \small
  \setlength{\tabcolsep}{2.6pt}
  \input{tables/tab_adult_values}
  \caption{Selected results from testing the UCI Adult dataset at $B = 19$: every value identified by the test for each demographic group.
    $n_R$ is the number of records with the value, $\hat p_R$ is its rate of group membership, and $\lcb_R$ is the lower bound of Equation~\eqref{eq:jeffreys}.
    The implied Bayes factor (BF) is computed from $\lcb_R$.}
  \label{tab:adult-values}
\end{table}

%% file: tables/tab_adult_values.tex

\begin{tabular}{lrcccccc}
    \toprule
                                                & \multicolumn{4}{c}{distributional proxy test} & \multicolumn{3}{c}{aggregate measures}                                                                    \\
    \cmidrule(lr){2-5} \cmidrule(lr){6-8}
    candidate value                             & $n_R$                                         & $\hat p_R$                             & $\lcb_R$       & implied BF & MI    & V $=|$corr$|=$ dcor & AUC  \\
    \midrule
    \multicolumn{8}{l}{\emph{sex $=$ Male} ($\pi{=}0.668$, $T{=}0.975$)}                                                                                                                                    \\
    \quad relationship $=$ Husband              & 19,716                                        & 1.000                                  & 1.000 & 2,502      & 0.225 & 0.58                & 0.80 \\
    \quad marital-status $=$ Married-civ-spouse & 22,379                                        & 0.889                                  & 0.886          & 3.8        & 0.101 & 0.43                & 0.73 \\
    \quad relationship $=$ Wife                 & 2,331                                         & 0.001                                  & 0.000          & 0.00       & 0.055 & 0.32                & 0.57 \\
    \midrule
    \multicolumn{8}{l}{\emph{sex $=$ Female} ($\pi{=}0.332$, $T{=}0.904$)}                                                                                                                                  \\
    \quad relationship $=$ Husband              & 19,716                                        & 0.000                                  & 0.000          & 0.00       & 0.225 & 0.58                & 0.80 \\
    \quad relationship $=$ Wife                 & 2,331                                         & 0.999                                  & 0.997 & 667        & 0.055 & 0.32                & 0.57 \\
    \quad occupation $=$ Priv-house-serv        & 242                                           & 0.942                                  & 0.913 & 21         & 0.004 & 0.09                & 0.50 \\
    \midrule
    \multicolumn{8}{l}{\emph{race $=$ White} ($\pi{=}0.855$, $T{=}0.991$)}                                                                                                                                  \\
    \quad (no values flagged)                   &                                               &                                        &                &            &       &                     &      \\
    \quad native-country $=$ United-States      & 43,832                                        & 0.878                                  & 0.876          & 1.2        & 0.015 & 0.19                & 0.58 \\
    \midrule
    \multicolumn{8}{l}{\emph{race $=$ Black} ($\pi{=}0.096$, $T{=}0.668$)}                                                                                                                                  \\
    \quad relationship $=$ Unmarried            & 5,125                                         & 0.223                                  & 0.214          & 2.6        & 0.009 & 0.15                & 0.57 \\
    \quad native-country $=$ Jamaica            & 106                                           & 0.925                                  & 0.874 & 65         & 0.004 & 0.13                & 0.50 \\
    \quad native-country $=$ Haiti              & 75                                            & 0.960                                  & 0.909 & 94         & 0.003 & 0.12                & 0.50 \\
    \midrule
    \multicolumn{8}{l}{\emph{race $=$ Asian-Pac.} ($\pi{=}0.031$, $T{=}0.379$)}                                                                                                                             \\
    \quad native-country $=$ United-States      & 43,832                                        & 0.010                                  & 0.009          & 0.28       & 0.035 & 0.36                & 0.81 \\
    \quad native-country $=$ Philippines        & 295                                           & 0.946                                  & 0.921 & 362        & 0.019 & 0.41                & 0.58 \\
    \quad native-country $=$ China              & 122                                           & 0.975                                  & 0.943 & 519        & 0.008 & 0.27                & 0.53 \\
    \quad native-country $=$ South              & 115                                           & 0.957                                  & 0.916 & 342        & 0.007 & 0.26                & 0.52 \\
    \quad native-country $=$ India              & 151                                           & 0.821                                  & 0.766 & 102        & 0.007 & 0.25                & 0.53 \\
    \quad native-country $=$ Vietnam            & 86                                            & 0.977                                  & 0.937 & 464        & 0.006 & 0.23                & 0.52 \\
    \quad native-country $=$ Taiwan             & 65                                            & 0.938                                  & 0.875 & 218        & 0.004 & 0.19                & 0.51 \\
    \quad native-country $=$ Japan              & 92                                            & 0.576                                  & 0.490 & 30         & 0.003 & 0.14                & 0.51 \\
    \quad native-country $=$ Hong               & 30                                            & 0.833                                  & 0.701 & 73         & 0.002 & 0.11                & 0.50 \\
    \quad native-country $=$ Thailand           & 30                                            & 0.833                                  & 0.701 & 73         & 0.002 & 0.11                & 0.50 \\
    \midrule
    \multicolumn{8}{l}{\emph{race $=$ Amer-Indian} ($\pi{=}0.010$, $T{=}0.156$)}                                                                                                                            \\
    \quad (no values flagged)                   &                                               &                                        &                &            &       &                     &      \\
    \quad education $=$ Bachelors               & 8,025                                         & 0.004                                  & 0.003          & 0.27       & 0.000 & 0.03                & 0.53 \\
    \midrule
    \multicolumn{8}{l}{\emph{race $=$ Other} ($\pi{=}0.008$, $T{=}0.137$)}                                                                                                                                  \\
    \quad native-country $=$ United-States      & 43,832                                        & 0.004                                  & 0.004          & 0.46       & 0.005 & 0.13                & 0.70 \\
    \quad native-country $=$ Dominican-Republic & 103                                           & 0.204                                  & 0.145 & 20         & 0.001 & 0.10                & 0.50 \\
    \quad native-country $=$ Ecuador            & 45                                            & 0.267                                  & 0.171 & 25         & 0.001 & 0.09                & 0.49 \\
    \bottomrule
\end{tabular}

%% file: sections/E_causal_proxy.tex
\section{The Federal Reserve Definition of a Proxy as a Causal Graph}\label{app:causal-proxy}

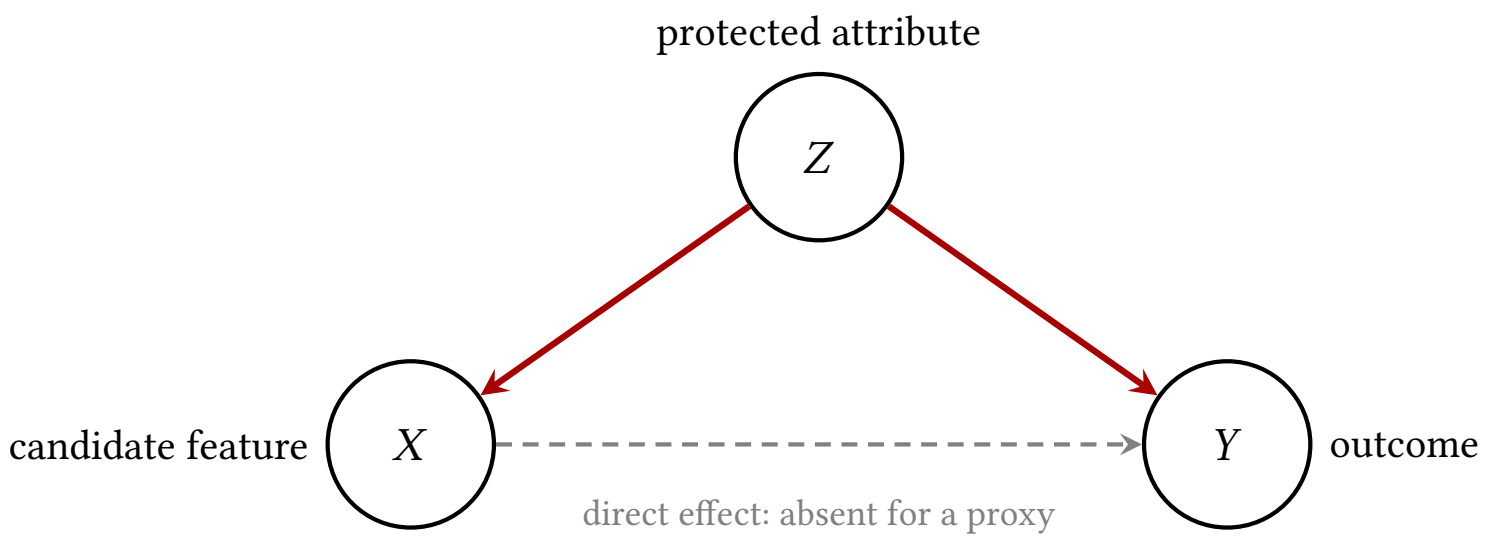
\begin{figure}[h]
  \centering
  \begin{tikzpicture}[>=stealth,
      var/.style={circle, draw, thick, minimum size=1.1cm, inner sep=1pt},
      lbl/.style={font=\small}]
    \node[var, label={[lbl]above:protected attribute}] (Z) at (2.7,1.9) {$Z$};
    \node[var, label={[lbl]left:candidate feature}]    (X) at (0,0)   {$X$};
    \node[var, label={[lbl]right:outcome}]             (Y) at (5.4,0) {$Y$};
    \draw[->, very thick, red!65!black] (Z) -- (X);
    \draw[->, very thick, red!65!black] (Z) -- (Y);
    \draw[->, thick, densely dashed, gray]
    (X) -- node[font=\footnotesize, below=7pt, gray] {direct effect: absent for a proxy} (Y);
  \end{tikzpicture}
  \Description{Causal graph with three nodes: protected attribute Z at the top, candidate feature X at the left, and outcome Y at the right. Solid arrows run from Z to X and from Z to Y. A dashed arrow from X to Y is labeled as the direct effect that is absent for a proxy.}
  \caption{A causal graph for the Federal Reserve's definition of a proxy. A candidate
    feature $X$ proxies for the protected attribute $Z$ when $X$ is associated
    with $Z$ and its association with the outcome $Y$ flows entirely through $Z$ (solid red).
    So in the proxy case, $X$ is independent of $Y$ given $Z$.
    When the dashed direct edge is present, $X$ carries predictive
    signal of $Y$ when conditioned on $Z$,
    so it is not deemed a proxy, regardless of the strength of association between $X$ and $Z$.}
  \label{fig:causal-proxy}
\end{figure}

Figure~\ref{fig:causal-proxy} shows the causal graph behind the definition of a proxy of \citet{avery2012credit}, which Section~\ref{sec:intro} cites.
The graph relates $X$ to the outcome $Y$ in the data.

